\documentclass[journal]{IEEEtran}
\usepackage{amsmath,amsfonts}
\usepackage{algorithmic}
\usepackage{algorithm}
\usepackage{array}
\usepackage[caption=false,font=normalsize,labelfont=sf,textfont=sf]{subfig}
\usepackage{textcomp}
\usepackage{stfloats}
\usepackage{url}
\usepackage{verbatim}
\usepackage{graphicx}
\usepackage{cite}

\usepackage{booktabs}
\usepackage{color}
\usepackage{colortbl}
\usepackage{array}
\usepackage{multirow}
\usepackage{pifont}
\usepackage{bm}
\usepackage{tikz}
\usepackage{xcolor}
\usepackage{makecell}
\usepackage{mathtools}
\usepackage{mathrsfs}
\definecolor{Gray}{gray}{0.93}
\definecolor{selfblue}{RGB}{0,255,255}

\begin{document}

\title{SSPA: Split-and-Synthesize Prompting with Gated Alignments for Multi-Label Image Recognition} 

\author{
	Hao Tan, 
	Zichang Tan,~\IEEEmembership{Member,~IEEE}, 
        Jun Li,
        Jun Wan,~\IEEEmembership{Senior Member,~IEEE}, \\
        Zhen Lei,~\IEEEmembership{Fellow,~IEEE},
        Stan Z. Li,~\IEEEmembership{Fellow,~IEEE}
\thanks{Hao Tan, Jun Li, Jun Wan and Zhen Lei are with the State Key Laboratory of Multimodal Artificial Intelligence Systems (MAIS), Institute of Automation Chinese Academy of Sciences (CASIA), Beijing 100190, China, and also with the School of Artificial Intelligence, University of Chinese Academy of Sciences (UCAS), Beijing 100049, China.
Jun Wan is also with the School of Computer Science and Engineering, Macau University of Science and Technology, Macau, China.
Zhen Lei is also with the Centre for Artificial Intelligence and Robotics, Hong Kong Institute of Science \& Innovation, Chinese Academy of Sciences, Hong Kong, China (e-mail: \{tanhao2023, lijun2021, jun.wan, zhen.lei\}@ia.ac.cn).}
\thanks{Zichang Tan is with the Sangfor Technologies Inc., and also with the Shenzhen Institute of Advanced Technology (SIAT), Chinese Academy of Sciences (e-mail: tanzichang@foxmail.com).}
\thanks{Stan Z. Li is with the Westlake University (e-mail: Stan.ZQ.Li@westlake.edu.cn).}
}

\maketitle

\begin{abstract}
  Multi-label image recognition is a fundamental task in computer vision. 
  Recently, Vision-Language Models (VLMs) have made notable advancements in this area. 
  However, previous methods fail to effectively leverage the rich knowledge in language models and often incorporate label semantics into visual features unidirectionally. 
  To overcome these problems, we propose a \textbf{S}plit-and-\textbf{S}ynthesize \textbf{P}rompting with Gated \textbf{A}lignments (SSPA) framework to amplify the potential of VLMs. 
  Specifically, we develop an in-context learning approach to associate the inherent knowledge from LLMs.
  Then we propose a novel Split-and-Synthesize Prompting (SSP) strategy to first model the generic knowledge and downstream label semantics individually and then aggregate them carefully through the quaternion network.
  Moreover, we present Gated Dual-Modal Alignments (GDMA) to bidirectionally interact visual and linguistic modalities while eliminating redundant cross-modal information, enabling more efficient region-level alignments.
  Rather than making the final prediction by a sharp manner in previous works, we propose a soft aggregator to jointly consider results from all image regions.
  With the help of flexible prompting and gated alignments, SSPA is generalizable to specific domains.
  Extensive experiments on nine datasets from three domains (i.e., natural, pedestrian attributes and remote sensing) demonstrate the state-of-the-art performance of SSPA.
  Further analyses verify the effectiveness of SSP and the interpretability of GDMA.
  The code will be made public.
\end{abstract}


\section{Introduction}
\label{sec:intro}
\IEEEPARstart{M}{ulti-label} recognition (MLR)~\cite{wang2016cnn, wang2017multi, chen2018order, chen2019learning, wang2020multi, ridnik2021asymmetric, chen2019multi, you2020cross, yazici2020orderless, ye2020attention, zhao2021m3tr, liu2021query2label, lanchantin2021general, zhu2022two, ridnik2023ml, zhu2023scene, li2023patchct, guo2023texts}
is a fundamental task in the field of computer vision,
where multiple labels are supposed to be recognized in a single image.
This ability to capture the diversity of visual content is paramount in applications like image tagging~\cite{huang2023tag2text, mamat2023enhancing}, human attribute recognition~\cite{tan2019attention, tan2020relation}, and recommendation systems~\cite{yang2015pinterest, jain2016extreme}.

With the rise of large-scale vision-language pre-training (VLP)~\cite{radford2021learning, jia2021scaling}, 
many approaches~\cite{chen2019learning, you2020cross, wang2020multi, zhao2021m3tr, zhu2022two, zhu2023scene, li2023patchct} began to leverage linguistic modality to mitigate the lack of semantic information from a single visual input.
Typically, these approaches involve extracting semantic knowledge from the language model and employing it as supplementary information to assist the visual model in learning better label representations.
Due to the extensive semantics embedded in the language model and the well-aligned cross-modal features learned by VLP, 
these methods have made remarkable progress in multi-label recognition. 

\begin{figure*}[!t]
    \centering
    \includegraphics[width=1.0\linewidth]{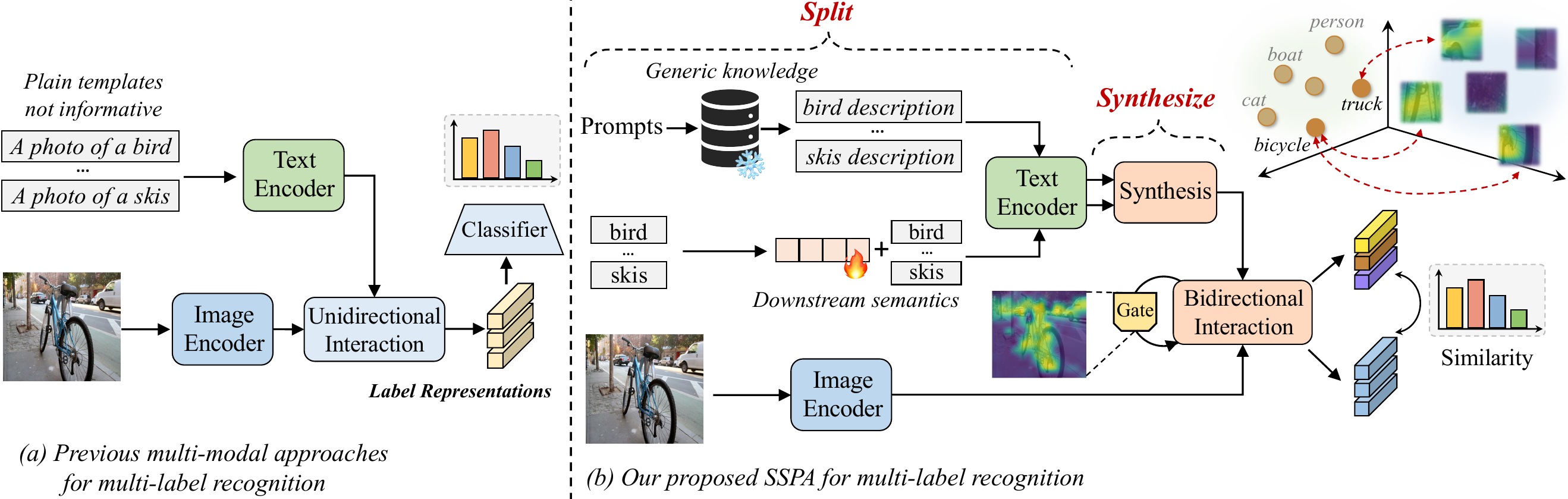}
    \caption{Paradigm comparison. (a) Previous methods~\cite{chen2019learning, you2020cross, wang2020multi, zhu2022two, zhu2023scene} adopt pure category names or plain templates to extract text features.
    Then unidirectional interaction is applied to aggregate label semantics and $C$ classifiers are trained for recognition. 
    (b) Our method employs novel split-and-synthesize prompting, where we extract generic knowledge from LLM and downstream semantics individually and then aggregate them in quaternion space.
    With the help of gated bidirectional interaction, SSPA can efficiently align the text features and regional features.
    No extra classifiers are required.}
	\label{fig:intro} 
\end{figure*}

However, the application of vision-language modeling in the domain of multi-label image recognition is still in its infancy. 
As summarized in Fig.~\ref{fig:intro}, there are mainly two limitations in existing works~\cite{chen2019learning, you2020cross, wang2020multi, zhu2022two, zhu2023scene, zhao2021m3tr, li2023patchct, abdelfattah2023cdul, xu2022dual}: 
\textbf{1)} Existing methods take pure category names~\cite{chen2019learning, you2020cross, wang2020multi, zhu2022two, zhu2023scene} or plain templates~\cite{li2023patchct, abdelfattah2023cdul} (e.g., ``\texttt{A photo of a \{category\}.}'') as the inputs of the text encoder, 
which lack sufficient knowledge acquisition for these powerful models and limit the fine-grained alignments between label semantics and spatial regions,
and thus often require explicit efforts to model the label relationships.
\textbf{2)} The linguistic modality is merely adopted as a supplement of semantic information for visual features\cite{zhu2023scene, you2020cross, zhao2021m3tr}, where $C$ additional classifiers ($C$ is the number of candidate labels) are required to learn the category centers, which remain static during inference, hindering the generalization abilities among different domains.
Besides, within the cross-modal interaction, different image patches (including background and negative areas) interact equivalently with label representations~\cite{xu2022dual}, introducing redundant information for multiple labels.
Such unshielded interaction significantly undermines the efficiency of cross-modal interaction.
For example, ADDS~\cite{xu2022dual} requires stacking six layers of cross-modal interaction to yield a satisfying performance, which is resource-consuming and inefficient.

To tackle the first problem, we propose a Split-and-Synthesize Prompting (SSP) strategy to first capture generic knowledge and downstream semantics individually and then aggregate them carefully to yield unified label representations. 
\textbf{In the split stage}, we first propose a Knowledge-Aware Prompting (KAP), where we query the inherent knowledge in Large Language Models (LLMs)~\cite{achiam2023gpt} through in-context learning~\cite{dong2022survey} to generate informative descriptions. 
The descriptions not only contain details that distinguish an object from other categories, but also provide underlying relationship information.
For instance, our generated sentence ``\texttt{Keyboard is a rectangular device with multiple keys, often found with mouses and computers.}'' 
describes the label relationship of ``\texttt{keyboard}'', ``\texttt{computer}'' and ``\texttt{mouse}''.
After fed into the text encoder, \textit{our method embeds the label relationships in the latent space rather than exhausting to model them explicitly}.
However, the KAP remains static and is not informative enough for the vastly changing visual context.
Therefore, inspired by recent success in prompt learning~\cite{zhou2022learning, zhou2022conditional}, we introduce Context-Aware Prompting (CAP), where we adopt learnable prompts to facilitate the learning of domain-relevant semantics.
Moreover, to capture visual-related label semantics, we incorporate a Dynamic Semantic Filtering (DSF) module in CAP to adaptively excavate the visual-related semantic cues.
Compared to KAP, CAP is capable of incorporating input-relevant label semantics, greatly facilitating the adaptation to specific domains. 
\textbf{In the synthesis stage}, 
we develop a simple yet general approach called Quaternion Semantic Modeling (QSM), which leverages quaternion network~\cite{parcollet2018quaternion} to aggregate generic knowledge and downstream semantics under the guidance of cross-modal orthogonal cues.
Since quaternion networks are proficient at relation modeling and exploration of inter-modal and intra-modal correlations~\cite{cao2024domain, wang2023quaternion}, the generic knowledge, downstream semantics and visual cues can be effectively synthesized within the quaternion latent space, generating unified representations for candidate labels.

For the second problem, we argue that simply enriching visual representations is unilateral and incomplete.
Since the label semantics might change as the visual scenes vary~\cite{zhu2023scene}, we consider to jointly rectify visual representations and adjust category centers.
To this end, we propose a Gated Dual-Modal Alignments (GDMA) module.
Specifically, we integrate visual information into the label embeddings, yielding context-aware label representations, and symmetrically integrate the label semantics into visual features, resulting in semantic-related visual representations. 
Different from common cross-modal interaction, we suggest that for multi-label recognition, some image regions (e.g., negative areas) might introduce redundant information for specific labels.
Directly aligning all image regions with label semantics might lead to noisy and subpar results.
Therefore, we propose a gated cross-modal attention mechanism to overcome the problem, where a gate vector is learned to suppress the redundant information and activate the useful cross-modal signals.
To facilitate the alignments between label semantics and regional features, we propose a soft aggregator to perform fine-grained alignments based on the importance score of each image patch.
In the prediction stage, unlike previous method~\cite{li2023patchct} that directly takes global visual feature to determine the existence of all labels, or CDUL~\cite{abdelfattah2023cdul} that employs a hard aggregator, our soft aggregator synthesize the information from different image patches in a smooth manner, which brings significant improvements.

To sum up the work, we propose a \textbf{S}plit-and-\textbf{S}ynthesize \textbf{P}rompting with Gated \textbf{A}lignments (SSPA) framework for multi-label image recognition. 
Our main contributions are:
\begin{itemize}
\item We propose a novel Split-and-Synthesize Prompting strategy, where generic knowledge and downstream semantics are modeled individually and then aggregated carefully in quaternion space. Leveraging quaternion network to aggregate distinct knowledge has not been explored before.
\item We present a Gated Dual-Modal Alignments module, where we first interact visual and label representations bidirectionally while eliminating redundant cross-modal signals, then we develop a soft aggregator to combine information from image regions for multi-label recognition. 
\item To the best of our knowledge, this is the first attempt to leverage inherent knowledge from LLMs to benefit multi-label image recognition through in-context learning, which further embeds the label relationships in the latent space rather than explicitly modeling them. The proposed prompting approach for LLMs is versatile and easy to transfer to other domains. 
\item With the help of flexible prompting and gated alignments, the proposed method is generalizable to specific domains. 
Extensive experiments on various datasets (including natural, pedestrian attributes and remote sensing) demonstrate the state-of-the-art performance of our method.
\end{itemize}

\section{Related Work}
\label{sec:related}
\subsection{General Multi-Label Classification}
\noindent Multi-label recognition serves as a fundamental task in the field of computer vision.
Most methods focus on the natural image domain, where the candidate labels are often common objects such as ``bicycle'' and ``dining table''.
Early approaches~\cite{wang2016cnn, yazici2020orderless, chen2018order, wang2017multi, ye2020attention, lanchantin2021general, tan2024vision} consider a single visual modality as input and focus on the modeling of label co-occurrence.
Some of them rely on the Recurrent Neural Network (RNN)~\cite{hochreiter1997long} and graph-based models~\cite{kipf2016semi}.
For example, Wang et al.~\cite{wang2016cnn} explore the semantic correlation among labels by cascading the RNN to the feature extractor.
Wang et al.~\cite{wang2017multi} utilize the LSTM to capture the dependencies among semantic regions.
Ye et al.~\cite{ye2020attention} propose a dynamic graph convolutional network (GCN) to model the content-aware relations for co-occurred categories.
Further works turn to visual attention for implicit relation mining.
For instance, Lanchantin et al.~\cite{lanchantin2021general} utilize a transformer encoder to explore the correlations among visual features and labels.
Zhu et al.~\cite{zhu2017learning} introduce self-attention to capture spatial relationships and then regularize the predictions.
With the help of attention mechanism, ML-Decoder~\cite{ridnik2023ml} proposes a versatile classification head consisting of cross-attention layers, while Q2L~\cite{liu2021query2label} shares similar ideas by stacking multiple transformer decoders to query the output logits.
Most recent works further enhance the performance from the perspective of knowledge distillation~\cite{rajeswar2022multi, yang2023multi} and loss improvements~\cite{kobayashi2023two}.

However, uni-modal methods lack label-related semantic information, which limits their generalization ability.
With the rise of language models such as BERT~\cite{devlin2018bert}, many approaches~\cite{chen2019learning, chen2019multi, wang2020multi, you2020cross, zhu2022two, zhao2021m3tr, zhu2023scene, li2023patchct} turn to leveraging linguistic modality to complement the semantic information.
Based on the extracted representations for candidate labels, these methods further focus on the interactions of different modalities.
You et al.~\cite{you2020cross} propose a cross-modality attention module to aggregate visual features and label embeddings, while additional classifiers are learned for predictions.
Similarly, Zhu et al.~\cite{zhu2022two} construct a two-stream transformer network to explore the textual-visual interactions and an MLP is trained for recognition.
Wang et al.~\cite{wang2020multi} superimpose multiple layers of graph to incrementally inject label semantics to feature learning.
Chen et al.~\cite{chen2019learning} and Zhu et al.~\cite{zhu2023scene} both insert semantic information to visual features unidirectionally through a low-rank bilinear pooling, while~\cite{zhu2023scene} further consider the scene-conditioned label co-occurrence.
Besides, some methods take advantage of VLMs to tackle open-vocabulary multi-label recognition, which is a more challenging task since the method needs to generalize to unseen targets.
DualCoOp~\cite{sun2022dualcoop} and DualCoOp++~\cite{hu2023dualcoop++} introduce minimal parameters to CLIP by learning coupled prompts. 
ADDS~\cite{xu2022dual} proposes a dual-modal decoder, while the interactions are unshielded and heavy, which requires stacking multiple layers to achieve better open-vocabulary performance.
MKT~\cite{he2023open} achieves this through a two-stream framework and feature distillation.
TAI~\cite{guo2023texts} proposes to learn prompts from text-only data, which could generalize to image recognition.

Besides natural image domain, multi-label recognition is also common in pedestrian and remote sensing images.
Pedestrian attribute recognition (PAR) aims to detect attributes (e.g., age and clothing) for images of pedestrians.
The prevalent approach is to explore attribute relationships.
For example, Tan et al.~\cite{tan2020relation} propose to learn attribute and contextual relations, respectively.
Recent methods also begin to utilize VLMs.
For example, Wang et al.~\cite{wang2023pedestrian} designs a region-aware prompting approach to transfer CLIP model, while the partition of regions is a strong prior for the pedestrian attributes domain and is not versatile to other tasks.
Multi-label remote sensing image classification is to recognize the aerial scenes (e.g., ``parking lot`` and ``forest'') in an aerial image.
GeRSP~\cite{huang2024generic} utilizes extra knowledge to boost the recognition in remote sensing domain.
RemoteCLIP~\cite{liu2024remoteclip} proposes a foundation vision-language model that shows exceptional generalization abilities for various remote sensing applications.

Different from previous multi-modal approaches, we explore the mutual interactions between visual and linguistic modalities, and further map the context-aware label representations into dynamic category centers, which greatly enhances the generalization.
Compared to the methods designed for specific MLR, our method is versatile to different domains.
A main challenge of MLR in specific domains is the semantic gap between target labels (e.g., human attributes or visual scenes) and natural objects,
while we mitigate this by resorting to LLMs for pertinent semantic extraction and applying gated mechanisms for robust alignments.

\subsection{Vision-Language Models}
\noindent Large-scale vision-language pre-training (VLP) has emerged as a powerful paradigm for a wide range of visual tasks~\cite{yan2023clip, lin2024mutual, liu2024cfpl}.
With a contrastive-based pre-training approach, vision-language models (VLMs) such as CLIP~\cite{radford2021learning} and ALIGN~\cite{jia2021scaling} learn a joint representation for visual and linguistic modalities, showing an encouraging ability for efficient transfer learning and zero-shot predictions.
More recently, there has been a growing interest in bridging pre-trained LLMs and vision foundation models to build VLM.
BLIP-2~\cite{li2023blip} achieves this by training an additional Q-former.
While MiniGPT-4~\cite{zhu2023minigpt} and LLaVA~\cite{liu2023visual} attain impressive multi-modal abilities by an extra linear projection.
The emergence of VLMs has benefited plenty of downstream tasks, e.g., 
person re-identification~\cite{yan2023clip}, action recognition~\cite{lin2024mutual} and face anti-spoofing~\cite{liu2024cfpl}.
Our method is built on CLIP, constructing an effective way to unleash the potential of VLMs in MLR.

\subsection{Prompt Learning}
Transferring the knowledge from the powerful VLMs to downstream tasks can be computationally expensive in the era of foundation models.
However, prompt learning which is initially introduced in the field of Natural Language Processing (NLP)~\cite{li2021prefix}, provides an efficient way for adapting VLMs.
CoOp~\cite{zhou2022learning} introduces a set of learnable prompts for textual input and yields remarkable few-shot performance.
CoCoOp~\cite{zhou2022conditional} instead adopts visual features to generate input-adaptive prompts.
The follow-up works~\cite{zhu2023prompt, khattak2023maple} further improve the generalization abilities.
For example, ProGrad~\cite{zhu2023prompt} achieves this by gradient correction,
and MaPLe~\cite{khattak2023maple} achieves it by introducing coupled prompts in both visual and text modalities.
Different from previous works that only use fixed prompts~\cite{li2023patchct, abdelfattah2023cdul} or learnable prompts~\cite{wang2023pedestrian, sun2022dualcoop} to adapt VLMs,
we for the first time propose a novel split-and-synthesize prompting strategy to facilitate the extraction of \textit{generic} and \textit{adaptive} label semantics concurrently.

\begin{figure*}[!t]
    \centering
    \includegraphics[width=0.96\linewidth]{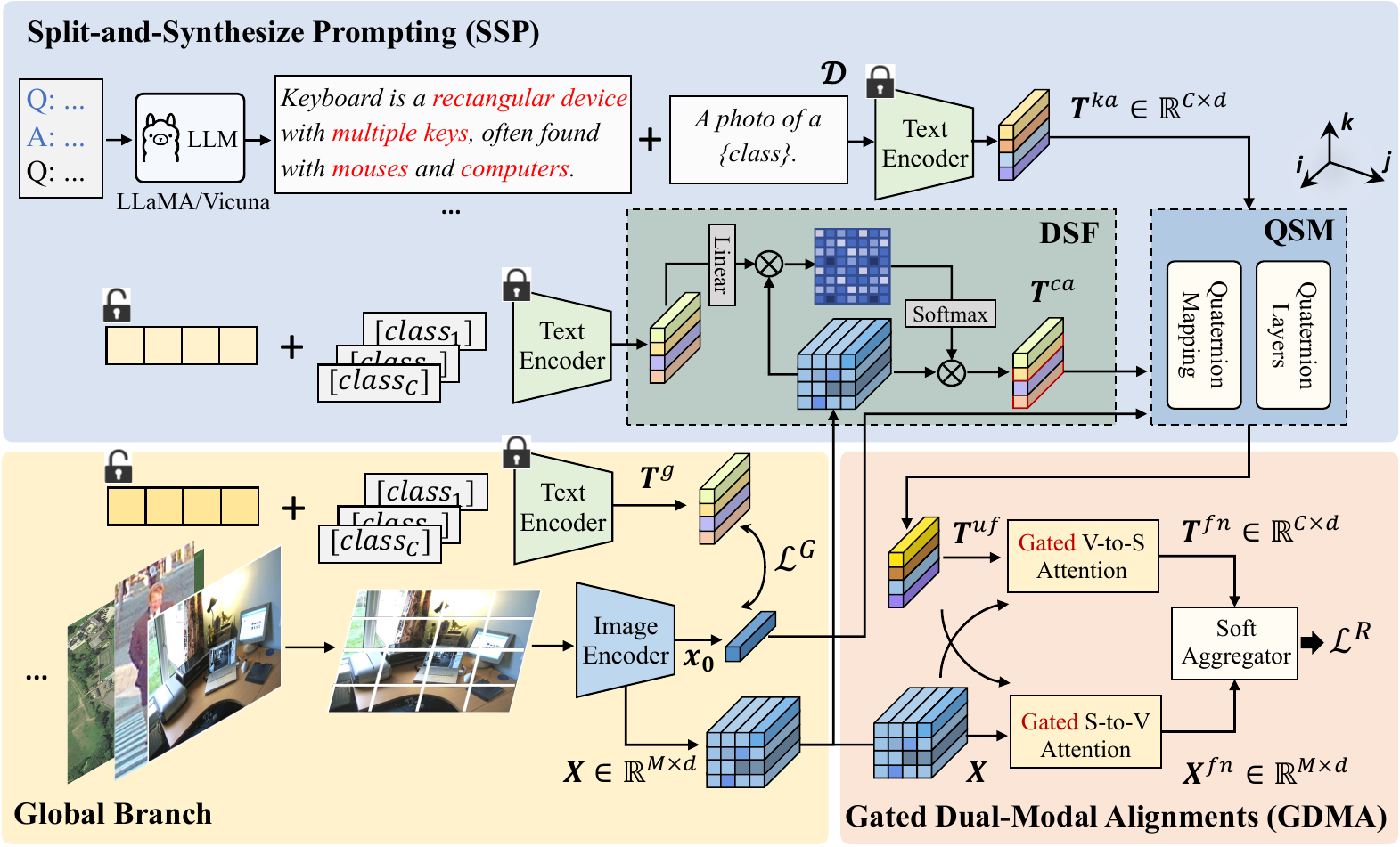}
 \caption{Overview of the proposed SSPA framework.
 The global branch directly compares global visual features with text features.
 The regional branch performs more fine-grained alignments between regional features and label semantics.
 We develop a Split-and-Synthesize Prompting (SSP) pipeline to get holistic label representations, where we concatenate LLM prompts with templates to get knowledge-aware text embeddings, and introduce learnable prompts and Dynamic Semantic Filtering (DSF) module to get context-aware text embeddings.
 Then we synthesize them through Quaternion Semantic Modeling (QSM) module.
 To mutually interact text embeddings and visual features while filtering out redundant cross-modal signals, we propose Gated Dual-Modal Alignments (GDMA), which efficiently aligns regional features with label semantics and achieves input-adaptive category centers during inference.
 The final scores are predicted based on our soft aggregator.
 }
	\label{fig:framework}
\end{figure*}

\subsection{Tool-Use of Large Language Models}
Large Language Models (LLMs) have demonstrated their comprehensive knowledge that can be beneficial in various NLP tasks, while this trend has extended to computer vision research.
Several studies have been proposed to investigate how LLMs can assist in downstream computer vision tasks.
CuPL~\cite{menon2022visual} leverages LLM-generated descriptors to enhance zero-shot classification.
Toubal et al.~\cite{toubal2024modeling} use LLM to enable subjective visual classification.
DVDet~\cite{jin2024llms} introduces interactive LLM to improve open-vocabulary object detection.
OPT2I~\cite{manas2024improving} applies LLM to revise the user prompts, improving the prompt-image consistency in text-to-image generation task.
SLD~\cite{wu2024self} utilizes LLM to post-rectify the generation results from diffusion models, reducing the error-prone generations when addressing complex user prompts.
In this work, we develop an efficient prompt template to leverage LLMs to generate pertinent descriptions for candidate labels, facilitating the generalization on different MLR domains.

\section{Proposed Method}
\label{sec:method}

\subsection{Preliminary}
\label{sec:pre}
\textbf{Notations.} 
\noindent For multi-label image recognition,
assume the input image $\boldsymbol{I}\in \mathbb{R}^{H\times W \times 3}$ is labeled with $C$ candidate categories, where $\boldsymbol{y}\in \mathbb{R}^{C}$ represents the multi-hot label vector and $\boldsymbol{y}_{j}=1$ means the input image contains the $j^{th}$ label and vice versa. 
For the input image $\boldsymbol{I}$,
we employ the pre-trained image encoder (e.g., ResNet~\cite{he2016deep} or ViT~\cite{dosovitskiy2020image}) from CLIP~\cite{radford2021learning} to extract visual features, denoted as $[\bm{x}_0, \bm{X}]=\texttt{image-encoder}(\bm{I})$, where $\boldsymbol{X}=[\bm{x}_1, ..., \bm{x}_M] \in \mathbb{R}^{M \times d}$ and $\bm{x}_0$ denotes the class token in ViT or pooled features in ResNet, which stands for the global visual feature.
$M$ indicates the number of patches, and $d$ is the feature dimension.

\noindent\textbf{Quaternion networks.}
A quaternion $Q$ is expressed in a hyper-complex space, which extends the real values into a four-dimensional space, known as the quaternion algebra $\mathbb{H}$:
\begin{equation}
    Q = r1 + x\textbf{i} + y\textbf{j} + z\textbf{k},\quad r, x, y, z \in \mathbb{R},
\end{equation}
where $r$ is the real part of $Q$ and $(\textbf{i}$, $\textbf{j}, \textbf{k})$ denote the orthogonal imaginary axes with $\textbf{i}^2 = \textbf{j}^2 = \textbf{k}^2 = \textbf{i}\textbf{j}\textbf{k}=-1$.
Typically, quaternion networks are useful for capturing complex correlations and interdependencies among diverse features and they contain embedded information that can be expressed by a real matrix~\cite{parcollet2018quaternion} as follows:
\begin{equation}
{Q}^{mat} = \left[ {\begin{array}{*{30}{c}}
r&{  - x  }&{  - y  }&{  - z}\\
x&{  r  }&{  - z  }&{  y  }\\
y&{  z  }&{  r  }&{  - x  }\\
z&{  - y  }&{  x  }&{  r  }
\end{array}} \right].
\end{equation}
Different from common operations in real-valued space, quaternion networks calculate on quaternions for all parameters, including inputs, outputs, weights and bias.
For the real-valued feature vectors, the corresponding quaternion can also be expressed in matrix form $\bm{Q} = \bm{r} + \textbf{i}\bm{x} + \textbf{j}\bm{y} + \textbf{k}\bm{z}$, where $\bm{r}, \bm{x}, \bm{y}$ and $\bm{z}$ are real-valued vectors.
We suggest using quaternion networks to establish multi-perspective relationships among diverse features~\cite{cao2024domain, wang2023quaternion}.
In this paper, we adopt quaternion networks to effectively aggregate generic knowledge, downstream semantics and visual context.

\noindent\textbf{Cross-Modal Attention.}
Cross-modal attention (CMA) takes different sources as input and is good at capturing cross-modal interactions.
Different from standard cross-attention\cite{vaswani2017attention}, we only introduce linear projection $\bm{W}_E$ for the \textit{query}.
Besides reducing complexity, removing linear mappings of \textit{key} and \textit{value} preserves the original semantics, while projecting query controls the transmission of cross-modal signals.
Suppose the inputs are denoted as $\boldsymbol{E}$ and $\boldsymbol{Z}$, the process is formulated as:
\begin{equation}
    \text{CMA}(\boldsymbol{E}, \boldsymbol{Z}) = \text{softmax}( \frac{(\boldsymbol{E}\bm{W}^E)\boldsymbol{Z}^{\mathsf{T}}}{\sqrt{d}})\boldsymbol{Z}.
\end{equation}

\subsection{Framework Overview}
As shown in Fig.~\ref{fig:framework}, the framework of SSPA is conceptually simple: the decision is based on both global and regional perspectives.
For global branch, we simply align the label representations (denoted as $\bm{T}^{g}$) with the global image feature $\bm{x}_0$, which largely preserves the knowledge of the original CLIP.
For regional alignment branch, since there are richer visual context, more fine-grained label semantics are supposed to improve the cross-modal alignments.
Therefore, we first use SSP to extract knowledge-aware and context-aware label semantics, which are then aggregated in the proposed QSM.
Then, the GDMA is applied to perform gated alignments between visual context and label semantics.
In the following sections, we focus on the details of the regional branch.

\subsection{Split-and-Synthesize Prompting}
\label{sec:ssp}
\begin{figure}[!t]
    \centering
    \includegraphics[width=1\linewidth]{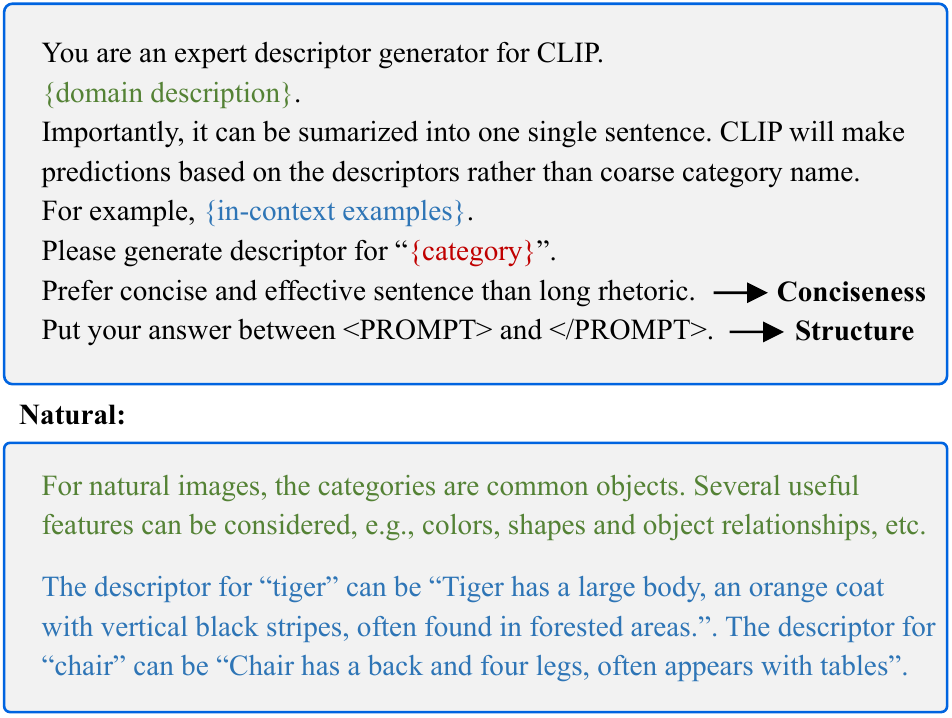}
    \caption{\textbf{Our prompts to LLM.} The text prompts the LLM to associate pertinent knowledge about shapes, sizes, colors and possible label relationships. Through domain description and in-context examples, LLM can be seamlessly linked to different domains. We also control the conciseness and structure of the answers to enable automatic processing.}
    \label{fig:llm}
\end{figure}

\noindent\textbf{Knowledge-Aware Prompting (KAP).}
For multi-label recognition, the essential factors are to distinguish objects from other similar categories and discover the label relationships.
We aim to leverage inherent knowledge in LLMs to assist the process.
However, naive prompts to LLMs result in descriptions of poor quality, for instance, ``\texttt{\underline{Airplane} is a powered flying vehicle used for transportation.}'', which is not useful and informative for recognition.
Instead, as shown in Fig.~\ref{fig:llm}, we develop an efficient prompt template to encourage the LLM to generate detailed descriptions (e.g., shapes, colors and sizes) with underlying label relationships, such as ``\texttt{\underline{Airplane} is a large aircraft with wings, engines, and jet or propeller engines, often seen in the sky or airports.}'',
which depicts notable features of airplane and relationships with ``\texttt{sky}'' and ``\texttt{airport}''.
By shifting the ``domain description'' and ``in-context examples'' parts in the prompts, we can activate the corresponding knowledge of LLMs and seamlessly transfer the model to specific domains.
After fed into the pre-trained text encoder, the generic knowledge and label relationships are embedded into the latent space.

Suppose the prompts are denoted as $\mathcal{P}$, we take the LLM to generate descriptions $\mathcal{D}^{LLM} = \text{LLM}(\mathcal{P})$.
Note that our method is flexible to the choice of LLM, and we use LLaMA3 in our experiments.
To encourage attention to the target object, we further concatenate a sentence to the $\mathcal{D}^{LLM}$.
The final description for the $j^{th}$ label is $\mathcal{D}_j=\text{``}\{\mathcal{D}^{LLM}_j\}.$ \texttt{A photo of a \{category$_j$\}.}'', which is then fed into text encoder:
\begin{equation}
    \bm{t}_j^{ka} = \text{text-encoder}(\mathcal{D}_j),\quad j=1,2,...,C,
\end{equation}
where $\bm{T}^{ka}=[\bm{t}_1^{ka}, \bm{t}_2^{ka},..., \bm{t}_C^{ka}] \in \mathbb{R}^{C\times d}$ is the knowledge-aware text embeddings.

\noindent\textbf{Context-Aware Prompting (CAP).}
The knowledge from KAP is static and agnostic to the input context.
To better capture the downstream knowledge, we introduce learnable prompts in CAP to facilitate the learning of task-relevant label semantics.
Inspired by CoOp~\cite{zhou2022learning}, we prepend $L$ prompt tokens to each label.
To reduce complexity, the prompt tokens are shared among all candidate labels.
The resulted input is denoted as $S_j=[\bm{p}^{1}][\bm{p}^{2}]...[\bm{p}^{L}][\boldsymbol{c}_j]$, 
where $\bm{p}^{l}\in \mathbb{R}^{d}$ is the $l^{th}$ learnable token, and $\bm{c}_j$ is the word embedding of the $j^{th}$ category name.
Then $S$ is fed into the text encoder to get the learnable text features $\bm{T}^{ln}$.

To further collect context-aware semantics, we propose a Dynamic Semantic Filtering (DSF) module.
Specifically, we compare each text feature in $\bm{T}^{ln}$ with visual region features $\bm{X}$ to determine the importance of visual context.
Then, the features for each label are extracted based on the importance weights.
Therefore, the resulted text features are filtered by the input visual context and are adaptive to varied visual scenes.
In practice, to preserve the CLIP extracted visual context, we only perform linear transformation to the text features, which results in a cross-modal attention layer.
The process of CAP can be summarized as follows:
\begin{equation}
\begin{aligned}
    \bm{t}_j^{ln} &= \text{text-encoder}(S_j),\quad j=1,2,...,C, \\
    \bm{\widehat{T}}^{ln} &= \text{CMA}(\bm{T}^{ln}, \bm{X}) + \bm{T}^{ln}, \\
    \bm{T}^{ca} &= \text{MLP}(\bm{\widehat{T}}^{ln}) + \bm{\widehat{T}}^{ln}, \\
\end{aligned}
\end{equation}
where $\bm{T}^{ln}=[\bm{t}_1^{ln}, \bm{t}_2^{ln},..., \bm{t}_C^{ln}] \in \mathbb{R}^{C\times d}$ denotes the learnable text embeddings and $\text{MLP}$ is a multi-layer perceptron.
$\bm{T}^{ca}=[\bm{t}_1^{ca}, \bm{t}_2^{ca},..., \bm{t}_C^{ca}] \in \mathbb{R}^{C\times d}$ represents the context-aware text embeddings.
For simplicity, we omit all the symbols for layernorm operations.

\noindent\textbf{Synthesis.}
Different from the conventional feature fusion methods that typically employ concatenation or summation, 
we propose a Quaternion Semantic Modeling (QSM) module to aggregate the knowledge-aware text embeddings $\bm{T}^{ka}$ and context-aware text embeddings $\bm{T}^{ca}$.
Furthermore, we take global visual feature $\bm{x}_0$ to guide the synthesis of generic knowledge and downstream semantics.
The simple combination of $\{\bm{T}^{ka}, \bm{T}^{ca}, \bm{x}_0\}$ yields a rudimentary multi-perspective feature.
Then we take a linear transformation $\bm{W}^Q$ to encode it into quaternion latent space:
\begin{equation}
    \bm{F}^{mp} = (\bm{T}^{ca} + \bm{T}^{ka} + \bm{x}_0)\bm{W}^Q,\quad \bm{F}^{mp} \in \mathbb{R}^{C\times d}.
\end{equation}
Then, we map $\bm{F}^{mp}$ into the quaternion by slicing up its real-valued features into four equidimensional vectors:
\begin{equation}
\begin{aligned}
    &\bm{Q}^{mp} = \bm{F}^{mp}_{s1} + \textbf{i}\bm{F}^{mp}_{s2} + \textbf{j}\bm{F}^{mp}_{s3} + \textbf{k}\bm{F}^{mp}_{s4},\\
    &\text{where} \quad \bm{F}^{mp}_{si} \in \mathbb{R}^{C\times \frac{d}{4}},\quad i=1,2,3,4.
\end{aligned}
\end{equation}
Such mapping does not introduce extra trainable layers, making our method simple and universal.
Suppose the quaternion weight matrix is $\bm{W} = \bm{W}^R + \textbf{i}\bm{W}^I + \textbf{j}\bm{W}^J + \textbf{k}\bm{W}^K$, where $\bm{W}^R, \bm{W}^I, \bm{W}^J$ and $\bm{W}^K$ are real-valued trainable matrix.
Performing the linear operation between $\bm{W}$ and $\bm{Q}^{mp}$ by using distributive property we get:
\begin{small}
\begin{equation}
    \left[\!
    \begin{matrix}
    \mathscr{R}(\bm{W}\!\otimes\!\bm{Q}^{mp}) \\
    \mathscr{I}(\bm{W}\!\otimes\!\bm{Q}^{mp}) \\
    \mathscr{J}(\bm{W}\!\otimes\!\bm{Q}^{mp}) \\
    \mathscr{K}(\bm{W}\!\otimes\!\bm{Q}^{mp}) \\
    \end{matrix}
    \!\right]
    \!=\!
    \left[\!
    \begin{matrix}
    \bm{W}^R\!&\!-\!\bm{W}^I\!&\!-\!\bm{W}^J\!&\!-\!\bm{W}^K \\
    \bm{W}^I\!&\!\bm{W}^R\!&\!-\!\bm{W}^K\!&\!\bm{W}^J \\
    \bm{W}^J\!&\!\bm{W}^K\!&\!\bm{W}^R\!&\!-\!\bm{W}^I \\
    \bm{W}^K\!&\!-\!\bm{W}^J\!&\!\bm{W}^I\!&\!\bm{W}^R \\
    \end{matrix}\!
    \right]\!
    \left[\!
    \begin{matrix}
    \bm{F}^{mp}_{s1} \\
    \bm{F}^{mp}_{s2} \\
    \bm{F}^{mp}_{s3} \\
    \bm{F}^{mp}_{s4} \\
    \end{matrix}
    \right]\!,
\end{equation}
\end{small}
where $\otimes$ is known as the Hamilton product~\cite{parcollet2018quaternion}. $\mathscr{R}(\cdot), \mathscr{I}(\cdot), \mathscr{J}(\cdot)$ and $\mathscr{K}(\cdot)$ are to obtain the features on the real or imaginary components. 
The quaternion linear operation is similar to a mixture of standard linear and channel-wise separable linear transformations.
Suppose the quaternion linear layer is denoted as $Q^{\psi}$:
\begin{equation}
    Q^{\psi}(\bm{h}) = \bm{W}\otimes \bm{h},
\end{equation}
where $\bm{h}$ is the input vector.
The quaternion layers used in this work are formulated as follows:
\begin{equation}
\begin{aligned}
    \bm{\widehat{F}}^{mp}&=\phi(Q^{\psi}_1(\bm{F}^{mp})),\\
    \bm{T}^{uf}&=\phi(Q^{\psi}_2(\bm{\widehat{F}}^{mp})),
\end{aligned}
\end{equation}
where $\phi$ is the ReLU activation.
By stacking two quaternion linear layers $Q^{\psi}_1$ and $Q^{\psi}_2$,
the interdependencies between generic knowledge and downstream semantics, and cross-modal signals are well modeled in the quaternion latent space, which generates holistic and unified label representations $\bm{T}^{uf}$.

\subsection{Gated Dual-Modal Alignments}
\noindent While previous methods perform unidirectional interaction between visual and linguistic modalities, we consider this as a mutually promoting process. 
However, the dense interaction across modalities would contain noisy and redundant signals, which are misleading for the recognition of specific categories.
Therefore, we introduce a gate mechanism into our cross-modal alignments, which performs bidirectional interactions while eliminating redundant signals.

\noindent\textbf{Gate mechanism.}
Suppose the cross-modal inputs are denoted as $\bm{P}$ and $\bm{U}$, respectively. Inspired by~\cite{kumar2020gated}, the gate operation is employed as follows:
\begin{equation}
\begin{aligned}
    f(\bm{P}, \bm{U})\!&=\!tanh([\bm{P}, \bm{U}, \bm{P}\!-\!\bm{U}, \bm{P}\odot \bm{U}]\bm{W}^f + \bm{b}^f), \\
    v(\bm{P}, \bm{U})\!&=\!\sigma([\bm{P}, \bm{U}, \bm{P}\!-\!\bm{U}, \bm{P}\odot \bm{U}]\bm{W}^g + \bm{b}^g), \\
    g(\bm{P}, \bm{U})\!&=\!v(\bm{P}, \bm{U})\odot f(\bm{P}, \bm{U}) + [1-v(\bm{P}, \bm{U})]\odot \bm{U},
\end{aligned}
\label{eq:gate}
\end{equation}
where $\{\bm{W}^f, \bm{W}^g, \bm{b}^f, \bm{b}^g\}$ are trainable parameters, $\odot$ denotes the Hadamard product, and $\sigma$ is sigmoid function. $f(\bm{P}, \bm{U})$ denotes the modulated input, and $v(\bm{P}, \bm{U})$ is a learned gate vector with each element controlling the influence of a corresponding pair of interactions.
$g(\bm{P}, \bm{U})$ denotes gated output.

\begin{figure}[!t]
    \centering
    \includegraphics[width=0.71\linewidth]{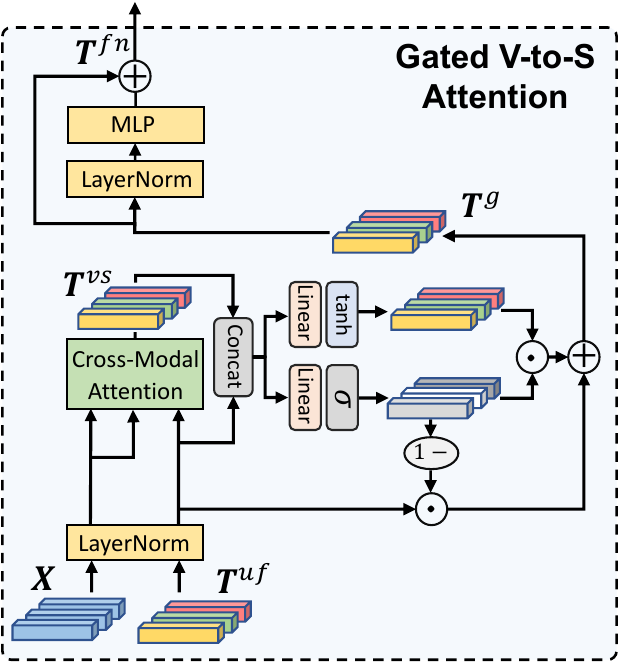}
    \caption{\textbf{The proposed gated visual-to-semantic attention.} The output of cross-modal attention is gated by the learned gate vector to filter out redundant signals. Gated semantic-to-visual attention is of symmetric structure.}
    \label{fig:gdma}
\end{figure}

\noindent\textbf{Gated dual-modal alignments.}
GDMA consists of two symmetric attention modules, i.e., gated visual-to-semantic attention and gated semantic-to-visual attention.
As shown in Fig.~\ref{fig:gdma}, the interactions are first fully explored by cross-modal attention, and then gated by Eq.~\ref{eq:gate}.
For gated visual-to-semantic attention, the process can be summarized as:
\begin{equation}
\begin{aligned}
    \bm{T}^{vs} &= \text{CMA}(\bm{T}^{uf}, \bm{X}), \\
    \bm{T}^{g} &= g(\bm{T}^{vs}, \bm{T}^{uf}), \\
    \bm{T}^{fn} &= \text{MLP}(\bm{T}^{g}) + \bm{T}^g,
\end{aligned}
\end{equation}
where $\bm{T}^{fn} \in \mathbb{R}^{C\times d}$ denotes the final label representations.
Similarly, for gated semantic-to-visual attention, the process is summarized as follows:
\begin{equation}
\begin{aligned}
    \bm{X}^{sv} &= \text{CMA}(\bm{X}, \bm{T}^{uf}), \\
    \bm{X}^{g} &= g(\bm{X}^{sv}, \bm{X}), \\
    \bm{X}^{fn} &= \text{MLP}(\bm{X}^{g}) + \bm{X}^g,
\end{aligned}
\end{equation}
where $\bm{X}^{fn} \in \mathbb{R}^{M\times d}$ is the final visual representations.
Since $\bm{T}^{fn}$ encapsulates both generic semantic knowledge and downstream visual context, it can seamlessly serve as the category centers for candidate labels, without additional learning of classifiers.
During inference, our method can simultaneously rectify the visual representations and adjust the category centers according to the specific inputs, yielding much better generalization performance.
   
\noindent \textbf{Soft aggregator.} To jointly consider information from different image regions and facilitate fine-grained alignments, we propose a soft aggregator. 
Specifically, the presence probability $p_{ij}^R$ of the $j^{th}$ label in the $i^{th}$ local patch is computed as:
\begin{equation}
    p_{ij}^R = \bm{X}_i^{fn} (\bm{T}_j^{fn})^{\mathsf{T}},\quad j=1,2,...,C.
\end{equation}
While background and negative areas might have low probabilities, we take the softmax to determine the importance of the $i^{th}$ local patch for the $j^{th}$ label:
\begin{equation}
    \gamma_{ij} = \frac{exp(p^R_{ij}/\tau)}{\sum_{k=1}^{M} exp(p^R_{kj}/\tau)},\quad i=1,2,...,M,
\end{equation}
where $\tau$ is a learnable temperature term.
The aggregated score of the $j^{th}$ label is calculated based on the weighted average of regional predictions, which is written as follows: 
\begin{equation}
    \widehat{p}^R_j = \sigma(\sum_{i=1}^{M} \gamma_{ij} p^R_{ij}),\quad j=1,2,...,C.
\end{equation}

\subsection{Training Objective}
In this work, we employ the Asymmetric Loss~\cite{ridnik2021asymmetric} for multi-label classification.
For the regional branch, the loss is calculated based on $\widehat{p}^R_j$, which is formulated as:
\begin{equation}
    \left.\mathcal{L}^{R}=\frac{1}{C}\sum_{j=1}^{C}\left\{\begin{aligned}(&1-\widehat{p}^R_{j})^{\gamma^+}\log \widehat{p}^R_{j},\quad \boldsymbol{y}_{j}=1,\\ 
    (&\widehat{p}^R_{j})^{\gamma^-}\log(1-\widehat{p}^R_{j}),\quad \boldsymbol{y}_{j}=0,
\end{aligned}\right.\right.
\label{eq:gamma}
\end{equation}
where $\gamma^+$ and $\gamma^-$ are asymmetric focusing parameters for positive and negative samples, respectively.

For the global branch, the predictions are generated by comparing the similarities between global visual feature $\bm{x}_{0}$ and text features $\bm{T}^g$, which is denoted as $p^G = \sigma(\bm{x}_0 (\bm{T}^{g})^{\mathsf{T}})$.
The global loss $\mathcal{L}^{G}$ is calculated by asymmetric loss similar to Eq.~\ref{eq:gamma}.
The final predicted score is the average from global and regional perspectives:
\begin{equation}
    p_j = \frac{p^G_j + \widehat{p}^R_j}{2},\quad j=1,2,...,C.
\end{equation}
The final training objective is formulated as:
\begin{equation}
    \mathcal{L}=\mathcal{L}^G + \lambda \mathcal{L}^R,
\label{eq:final_loss}
\end{equation}
where $\lambda$ is to control the importance of regional decisions.

\section{Experiments}
\label{sec:exp}
\subsection{Datasets and Metrics}
\noindent \textbf{Datasets.}
We conduct experiments on three popular benchmarks in MLR, i.e., MS-COCO~\cite{lin2014microsoft}, VOC 2007~\cite{everingham2010pascal} and NUS-WIDE~\cite{chua2009nus}.
\textbf{MS-COCO}~\cite{lin2014microsoft} contains 80 common categories and we use official train2014 (82K images) and val2014 (40K images) splits for training and testing.
\textbf{VOC 2007}~\cite{everingham2010pascal} contains 20 object categories and we use the official trainval (5K images) and test (5K images) splits for training and testing.
\textbf{NUS-WIDE}~\cite{chua2009nus} is more noisy, which contains 81 common categories and we use the train (12K images) and test (8K images) splits for training and testing.

\noindent Our method can be applied to other downstream domains, and we also evaluate on additional datasets from pedestrian attribute recognition (including PA100K~\cite{liu2017hydraplus}, RAPv1~\cite{li2018richly} and PETA~\cite{deng2014pedestrian}) and remote sensing image classification (including MultiScene~\cite{hua2021multiscene}, MLRSNet~\cite{qi2020mlrsnet} and AID~\cite{xia2017aid}).
\textbf{PA100K}~\cite{liu2017hydraplus} contains 26 pedestrian attributes and we take the official train (80K images) and test (10K images) splits for training and test.
\textbf{RAPv1}~\cite{li2018richly} contains 51 pedestrian attributes and we take the official train (30K images) and test (8K images) splits for training and test.
\textbf{PETA}~\cite{deng2014pedestrian} contains 35 pedestrian attributes and we take the official train (9K images) and test (7K images) splits for training and test.
\textbf{MultiScene}~\cite{hua2021multiscene} contains 36 aerial scenes and we take the ``clean'' set which contains 7K images for training and 7K images for testing.
\textbf{MLRSNet}~\cite{qi2020mlrsnet} is a large-scale remote sensing dataset which contains 60 aerial scenes and we take the official train (80K images) and test (20K images) splits for training and test.
\textbf{AID}~\cite{xia2017aid} contains 17 aerial scenes and we take the official train (2K images) and test (0.6K images) splits for training and test.

\noindent\textbf{Evaluation Metrics.}
For natural and remote sensing datasets, the mean average precision (mAP) is reported to evaluate the overall performance.
Following~\cite{wang2016cnn, chen2019multi, zhu2023scene}, we also report Class-wise Precision (CP), Recall (CR), F1 (CF1), and the average Overall Precision (OP), Recall (OR), F1 (OF1).
Note that ``CF1'' and ``OF1'' are more informative since Precision and Recall vary with the threshold.
To fairly compare with state-of-the-art, we further report top-3 results.
For pedestrian attribute recognition datasets, following previous work~\cite{li2018richly}, we adopt mean Accuracy (mA) as label-based metrics, Accuracy, Precision, Recall and F1-score as instance-based metrics.

\begin{table*}[t]
    \centering
    \caption{Comparison (\%) to state-of-the-art methods on MS-COCO. Results with different backbone and input resolutions are reported. mAP, OF1, and CF1 are primary metrics (highlighted in \textcolor{red}{\textbf{red}}) as the others may be affected by the threshold.}
    \scalebox{0.9}{
        \begin{tabular}{p{60pt}<{\raggedright}p{38pt}<{\centering}p{38pt}<{\centering}|p{18pt}<{\centering}p{16pt}<{\centering}p{16pt}<{\centering}p{16pt}<{\centering}p{16pt}<{\centering}p{16pt}<{\centering}p{16pt}<{\centering}|p{16pt}<{\centering}p{16pt}<{\centering}p{16pt}<{\centering}p{16pt}<{\centering}p{16pt}<{\centering}p{15pt}<{\centering}}
        \multirow{2}{*}{\hspace{-2pt}Method} & \multirow{2}{*}{Backbone} & \multirow{2}{*}{Resolution} & \multirow{2}{*}{mAP} & \multicolumn{6}{c|}{ALL} & \multicolumn{6}{c}{Top-3}\\
        & & & & CP & CR & CF1 & OP & OR & OF1 & CP & CR & CF1 & OP & OR & OF1 \\
        \cline{1-16}
        \rule{0pt}{7pt}\hspace{-2pt}ML-GCN~\cite{chen2019multi} & ResNet101 & (448, 448) & 83.0 & 85.1 & 72.0 & 78.0 & 85.8 & 75.4 & 80.3 & 89.2 & 64.1 & 74.6 & 90.5 & 66.5 & 76.7\\
        \hspace{-2pt}CMA~\cite{you2020cross} & ResNet101 & (448, 448) & 83.4 & 82.1 & 73.1 & 77.3 & 83.7 & 76.3 & 79.9 & 87.2 & 64.6 & 74.2 & 89.1 & 66.7 & 76.3\\
        \hspace{-2pt}TSGCN~\cite{xu2020joint} & ResNet101 & (448, 448) & 83.5 & 81.5 & 72.3 & 76.7 & 84.9 & 75.3 & 79.8 & 84.1 & 67.1 & 74.6 & 89.5 & 69.3 & 69.3\\
        \hspace{-2pt}CSRA~\cite{zhu2021residual} & ResNet101 & (448, 448) & 83.5 & 84.1 & 72.5 & 77.9 & 85.6 & 75.7 & 80.3 & 88.5 & 64.2 & 74.4 & 90.4 & 66.4 & 76.5\\
        \hspace{-2pt}ASL~\cite{ridnik2021asymmetric} & ResNet101 & (448, 448) & 85.0 & - & - & 80.3 & - & - & 82.3 & - & - & - & - & - & -\\
        \hspace{-2pt}TDRL~\cite{zhao2021transformer} & ResNet101 & (448, 448) & 84.6 & 86.0 & 73.1 & 79.0 & 86.6 & 76.4 & 81.2 & 89.9 & 64.4 & 75.0 & 91.2 & 67.0 & 77.2 \\
        \hspace{-2pt}Q2L-R101~\cite{liu2021query2label} & ResNet101 & (448, 448) & 84.9 & 84.8 & 74.5 & 79.3 & 86.6 & 76.9 & 81.5 & 78.0 & 69.1 & 73.3 & 80.7 & 70.6 & 75.4\\
        \hspace{-2pt}SALGL~\cite{zhu2023scene} & ResNet101 & (448, 448) & 85.8 & \textbf{87.2} & 74.5 & 80.4 & \textbf{87.8} & 77.6 & 82.4 & \textbf{90.4} & 65.7 & 76.1 & \textbf{91.9} & 67.9 & 78.1\\
        \hspace{-2pt}ML-Decoder~\cite{ridnik2023ml} & ResNet101 & (448, 448) & 86.6 & - & - & - & - & - & - & - & - & - & - & - & -\\
        \hspace{-2pt}DualCoOp~\cite{sun2022dualcoop} & ResNet101 & (448, 448) & 85.3 & - & - & - & - & - & - & - & - & - & - & - & -\\
        \rowcolor{Gray}\hspace{-2pt}\textbf{SSPA} & ResNet101 & (448, 448) & \textcolor{red}{\textbf{88.7}} & 83.1 & \textbf{82.8} & \textcolor{red}{\textbf{83.0}} & 83.4 & \textbf{85.9} & \textcolor{red}{\textbf{84.6}} & 89.1 & \textbf{69.3} & \textcolor{red}{\textbf{78.0}} & 90.8 & \textbf{71.4} & \textcolor{red}{\textbf{79.9}} \\
        \cline{1-16}
        \hspace{-2pt}\rule{0pt}{7pt}SSGRL~\cite{chen2019learning} & ResNet101 & (576, 576) & 83.6 & \textbf{89.5} & 68.3 & 76.9 & \textbf{91.2} & 70.7 & 79.3 & \textbf{91.9} & 62.1 & 73.0 & \textbf{93.6} & 64.2 & 76.0\\
        \hspace{-2pt}C-Tran~\cite{lanchantin2021general} & ResNet101 & (576, 576) & 85.1 & 86.3 & 74.3 & 79.9 & 87.7 & 76.5 & 81.7 & 90.1 & 65.7 & 76.0 & 92.1 & 71.4 & 77.6\\
        \hspace{-2pt}ADD-GCN~\cite{ye2020attention} & ResNet101 & (576, 576) & 85.2 & 84.7 & 75.9 & 80.1 & 84.9 & 79.4 & 82.0 & 88.8 & 66.2 & 75.8 & 90.3 & 68.5 & 77.9\\
        \hspace{-2pt}TDRL~\cite{zhao2021transformer} & ResNet101 & (576, 576) & 86.0 & 87.0 & 74.7 & 80.1 & 87.5 & 77.9 & 82.4 & 90.7 & 65.6 & 76.2 & 91.9 & 68.0 & 78.1 \\
        \hspace{-2pt}Q2L-R101~\cite{liu2021query2label} & ResNet101 & (576, 576) & 86.5 & 85.8 & 76.7 & 81.0 & 87.0 & 78.9 & 82.8 & 90.4 & 66.3 & 76.5 & 92.4 & 67.9 & 78.3\\
        \hspace{-2pt}SALGL~\cite{zhu2023scene} & ResNet101 & (576, 576) & 87.3 & 87.8 & 76.8 & 81.9 & 88.1 & 79.5 & 83.6 & 91.1 & 66.9 & 77.2 & 92.4 & 69.0 & 79.0 \\
        \rowcolor{Gray}\hspace{-2pt}\textbf{SSPA} & ResNet101 & (576, 576) & \textcolor{red}{\textbf{89.3}} & 83.2 & \textbf{83.6} & \textcolor{red}{\textbf{83.4}} & 83.6 & \textbf{86.5} & \textcolor{red}{\textbf{85.0}} & 88.7 & \textbf{69.9} & \textcolor{red}{\textbf{78.2}} & 91.2 & \textbf{71.6} & \textcolor{red}{\textbf{80.2}} \\
        \cline{1-16}
        \hspace{-2pt}\rule{0pt}{7pt}M3TR~\cite{zhao2021m3tr} & ViT-B/16 & (448, 448) & 87.5 & \textbf{88.4} & 77.2 & 82.5 & \textbf{88.3} & 79.8 & 83.8 & \textbf{91.9} & 68.1 & 78.2 & \textbf{92.6} & 69.6 & 79.4\\
        \hspace{-2pt}PatchCT~\cite{li2023patchct} & ViT-B/16 & (448, 448) & 88.3 & 83.3 & 82.3 & 82.6 & 84.2 & 83.7 & 83.8 & 90.7 & 69.7 & 78.8 & 90.3 & 70.8 & 79.8 \\
        \rowcolor{Gray}\hspace{-2pt}\textbf{SSPA} & ViT-B/16 & (448, 448) & \textcolor{red}{\textbf{90.1}} & 84.2 & \textbf{84.7} & \textcolor{red}{\textbf{84.5}} & 83.8 & \textbf{87.7} & \textcolor{red}{\textbf{85.7}} & 90.8 & \textbf{71.5} & \textcolor{red}{\textbf{80.0}} & 92.1 & \textbf{72.4} & \textcolor{red}{\textbf{81.1}}\\
        \end{tabular}
    }
    \label{tab:main_coco}
\end{table*}

\begin{table*}[t]
    \caption{Comparison (\%) to state-of-the-art methods on Pascal VOC 2007 in terms of class-wise average precision (AP) and mean average precision (mAP). \dag~indicates the ViT-B/16 backbone is used. We use abbreviations for some category names.}
    \label{tab:main_voc}
    \centering
    \scalebox{0.84}{
        \begin{tabular}{p{50pt}<{\raggedright}|p{14pt}<{\centering}p{14pt}<{\centering}p{14pt}<{\centering}p{14pt}<{\centering}p{14pt}<{\centering}p{14pt}<{\centering}p{14pt}<{\centering}p{14pt}<{\centering}p{14pt}<{\centering}p{14pt}<{\centering}p{14pt}<{\centering}p{14pt}<{\centering}p{14pt}<{\centering}p{14pt}<{\centering}p{14pt}<{\centering}p{14pt}<{\centering}p{14pt}<{\centering}p{14pt}<{\centering}p{14pt}<{\centering}p{14pt}<{\centering}|p{16pt}<{\centering}}
        \hspace{-2pt}Method & aero & bike & bird & boat & bottle & bus & car & cat & chair & cow & table & dog & horse & mot & psn & plt & shp & sofa & train & tv & mAP\\
        \cline{1-22}
        \rule{0pt}{7pt}\hspace{-2pt}SSGRL~\cite{chen2019learning} & 99.5 & 97.1 & 97.6 & 97.8 & 82.6 & 94.8 & 96.7 & 98.1 & 78.0 & 97.0 & 85.6 & 97.8 & 98.3 & 96.4 & 98.8 & 84.9 & 96.5 & 79.8 & 98.4 & 92.8 & 93.4\\
        \hspace{-2pt}ML-GCN~\cite{chen2019multi} & 99.5 & 98.5 & \textbf{98.6} & 98.1 & 80.8 & 94.6 & 97.2 & 98.2 & 82.3 & 95.7 & 86.4 & 98.2 & 98.4 & 96.7 & 99.0 & 84.7 & 96.7 & 84.3 & 98.9 & 93.7 & 94.0\\
        \hspace{-2pt}TSGCN~\cite{xu2020joint} & 98.9 & 98.5 & 96.8 & 97.3 & \textbf{87.5} & 94.2 & 97.4 & 97.7 & 84.1 & 92.6 & 89.3 & 98.4 & 98.0 & 96.1 & 98.7 & 84.9 & 96.6 & 87.2 & 98.4 & 93.7 & 94.3\\
        \hspace{-2pt}ASL~\cite{ridnik2021asymmetric} & - & - & - & - & - & - & - & - & - & - & - & - & - & - & - & - & - & - & - & - & 94.4\\
        \hspace{-2pt}CSRA~\cite{zhu2021residual} & 99.9 & 98.4 & 98.1 & \textbf{98.9} & 82.2 & 95.3 & 97.8 & 97.9 & 84.6 & 94.8 & \textbf{90.8} & 98.1 & 97.6 & 96.2 & 99.1 & 86.4 & 95.9 & 88.3 & 98.9 & 94.4 & 94.7\\
        \hspace{-2pt}SALGL~\cite{zhu2023scene} & 99.9 & \textbf{98.8} & 98.3 & 98.2 & 81.6 & 96.5 & \textbf{98.1} & 97.8 & 85.2 & 97.0 & 89.6 & 98.5 & 98.7 & 97.1 & 99.2 & 86.9 & 96.4 & \textbf{89.9} & 99.5 & 95.2 & 95.1\\
        \rowcolor{Gray} \hspace{-2pt}\textbf{SSPA} & \textbf{100.0} & 98.2 & 98.3 & 98.8 & 86.2 & \textbf{97.4} & \textbf{98.1} & \textbf{99.2} & \textbf{85.6} & \textbf{98.0} & 90.5 & \textbf{99.3} & \textbf{99.2} & \textbf{98.2} & \textbf{99.4} & \textbf{88.5} & \textbf{97.9} & 88.0 & \textbf{99.6} & \textbf{93.8} & \textbf{95.7} \\
        \cline{1-22}
        \hspace{-2pt}\rule{0pt}{7pt}Q2L-TRL~\cite{liu2021query2label} & 99.9 & 98.9 & 99.0 & 98.4 & 87.7 & 98.6 & 98.8 & 99.1 & 84.5 & 98.3 & 89.2 & 99.2 & 99.2 & \textbf{99.2} & 99.3 & 90.2 & 98.8 & 88.3 & 99.5 & 95.5 & 96.1\\
        \hspace{-2pt}M3TR\dag~\cite{zhao2021m3tr} & 99.9 & 99.3 & 99.1 & 99.1 & 84.0 & 97.6 & 98.0 & 99.0 & 85.9 & 99.4 & 93.9 & \textbf{99.5} & 99.4 & 98.5 & 99.2 & 90.3 & 99.7 & 91.6 & \textbf{99.8} & 96.0 & 96.5\\
        \hspace{-2pt}PatchCT$^{\dag}$~\cite{li2023patchct} & \textbf{100.0} & 99.4 & 98.8 & 99.3 & 87.2 & 98.6 & 98.8 & 99.2 & 87.2 & 99.0 & \textbf{95.5} & 99.4 & \textbf{99.7} & 98.9 & 99.1 & 91.8 & 99.5 & \textbf{94.5} & 99.5 & 96.3 & 97.1\\
        \rowcolor{Gray}\hspace{-2pt}\textbf{SSPA}$^{\dag}$ & \textbf{100.0} & \textbf{99.5} & \textbf{99.1} & \textbf{99.4} & \textbf{92.1} & \textbf{99.5} & \textbf{99.1} & \textbf{99.6} & \textbf{90.6} & \textbf{99.5} & 93.5 & \textbf{99.5} & 99.2 & 99.1 & \textbf{99.5} & \textbf{92.1} & \textbf{99.9} & 91.4 & 99.4 & \textbf{97.4} & \textbf{97.5} \\
        \end{tabular}
    }
\end{table*}

\subsection{Implementation Details}
\noindent Our method is built on CLIP~\cite{radford2021learning}.
The text encoder remains frozen.
The number of learnable prompt tokens $L$ is set to $4$.
$\gamma^{+}$ and $\gamma^{-}$ are set as $0$ and $2$, respectively.
$\lambda$ is set as 1.
For fair comparisons, the input images are resized to $448\times 448$ for natural image datasets, and $224\times 224$ for pedestrian attributes and remote sensing datasets.
The network is trained for $30$ epochs using AdamW~\cite{loshchilov2017decoupled} optimizer with a batch size of $32$.
The learning rate is set as $0.0001$ and decays with cosine policy.
Following previous works~\cite{ridnik2021asymmetric, zhu2023scene}, we apply exponential moving average with a decay of 0.9997.
We also implement CLIP-FT as a strong baseline, which fine-tunes all parameters including both image and text encoders.

\subsection{Comparison with State-of-the-art}
\noindent \textbf{Natural image MLR.}
The comparisons on MS-COCO, PASCAL VOC 2007, and NUS-WIDE are shown in Table~\ref{tab:main_coco}, Table~\ref{tab:main_voc} and Table~\ref{tab:main_nus}, respectively.
SSPA achieves state-of-the-art performance across various backbones and resolutions on all datasets, surpassing other methods with a decent margin.
On the MS-COCO, compared with multi-modal method SALGL~\cite{zhu2023scene}, our method exhibits considerable performance gains, exceeding them by 2.9\% mAP, which suggests the superiority of fully exploiting linguistic modality.
Compared with ML-GCN~\cite{chen2019multi} that also maps label representations into category centers while neglecting the visual context, SSPA achieves 5.7\% gains in mAP, demonstrating the effectiveness of learning input-adaptive category centers.
Moreover, our method outperforms all other methods on the resolution of $576\times 576$ and ViT-B/16 backbone, surpassing previous SOTA by 2.0\% and 1.8\% mAP respectively.
On the NUS-WIDE, our method surpasses all other methods on ResNet101 and ViT-B/16 backbones, achieving 67.7\% and 69.9\% mAP, respectively, which demonstrates the robustness of SSPA when addressing the noisy real-world images.
On the PASCAL VOC 2007, SSPA also outperforms all other methods.
With ViT-B/16 backbone, the AP on all 20 categories exceeds 90.6\%, achieving 97.5\% mAP,
which demonstrates the effectiveness of our method in handling objects of distinct sizes and semantics.
The experimental results clearly confirm the superiority of our proposed SSPA, 
and also show good generalizability to different network architectures.

\noindent \textbf{Pedestrian attribute recognition.}
As shown in Table~\ref{tab:main_par}, SSPA demonstrates superior performance compared to those specialized PAR approaches.
As for F1, our method achieves the best results on all datasets and all backbone settings, e.g., surpassing SOTA method by 0.9\%/2.2\% on RAPv1 using ResNet50/ViT-B.
As for mA, our method also achieves considerable improvements, e.g., outperforming SOTA method by 1.1\% on PA100K using ViT-B.
It is worth noting that compared to CLIP-FT that fine-tunes all parameters (including text and image encoders) and PromptPAR~\cite{wang2023pedestrian} that has further specialized designing, our method exhibits notable advantages on RAPv1 and PETA, which demonstrates the effectiveness and versatility of our usage of CLIP.

\noindent \textbf{Remote sensing image MLR.}
We fine-tune the most recent method~\cite{liu2024remoteclip} in the field of remote sensing recognition.
RemoteCLIP~\cite{liu2024remoteclip} pre-trains a CLIP-style model specialized in remote sensing tasks, which we further fine-tune on the MLR datasets.
As presented in Table~\ref{tab:main_rs}, 
SSPA achieves superior performance on all datasets, surpassing RemoteCLIP-FT by 1.2\%/1.3\% mAP on MultiScene using ResNet50/ViT-B.

\begin{figure*}[t]
    \centering
    \begin{minipage}[t]{0.25\textwidth}
    \centering
    \includegraphics[width=1\linewidth]{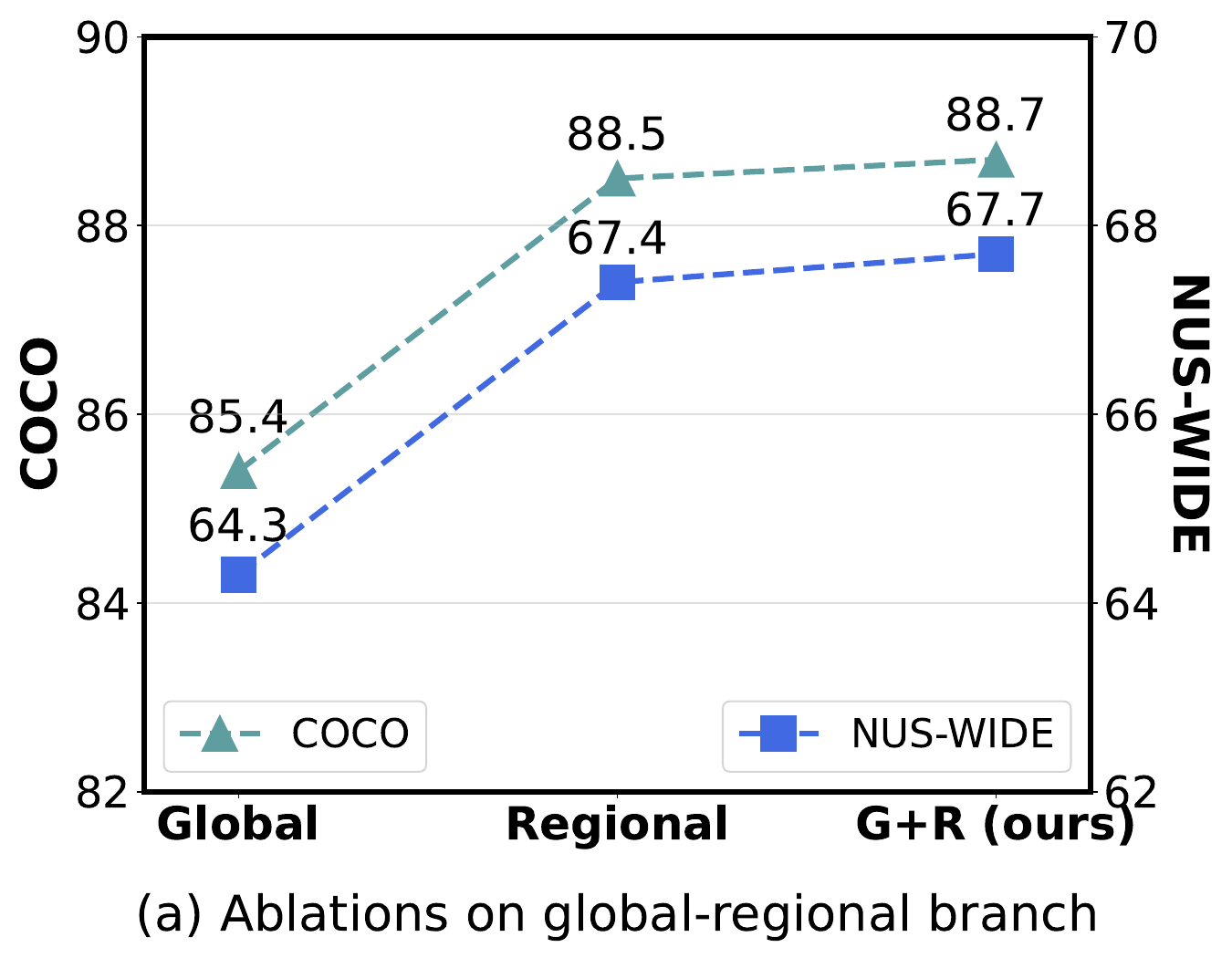}
    \end{minipage}
    \hspace{-3mm}
    \begin{minipage}[t]{0.25\textwidth}
    \centering
    \includegraphics[width=1\linewidth]{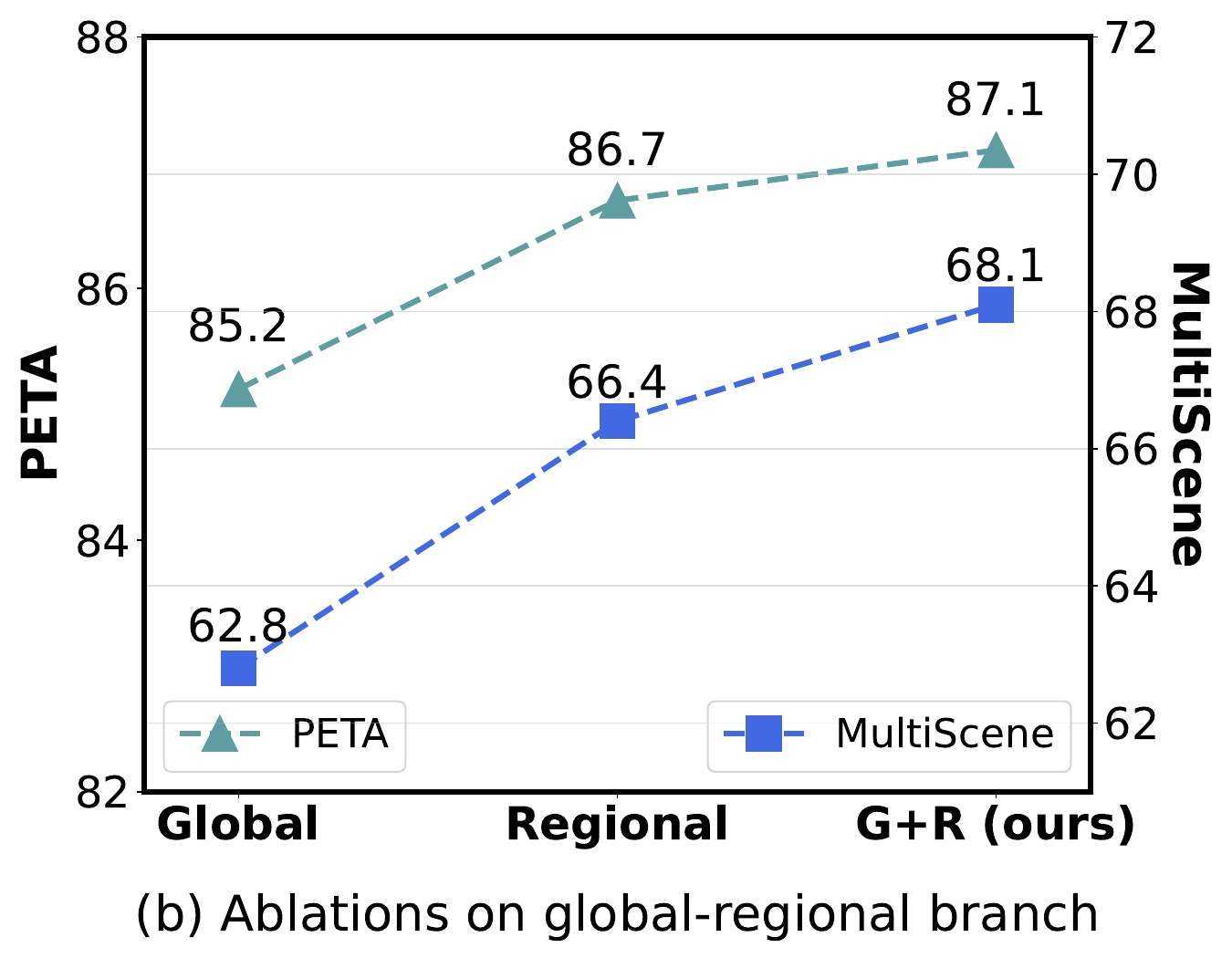}
    \end{minipage}
    \hspace{-3mm}
    \begin{minipage}[t]{0.25\textwidth}
    \centering
    \includegraphics[width=1\linewidth]{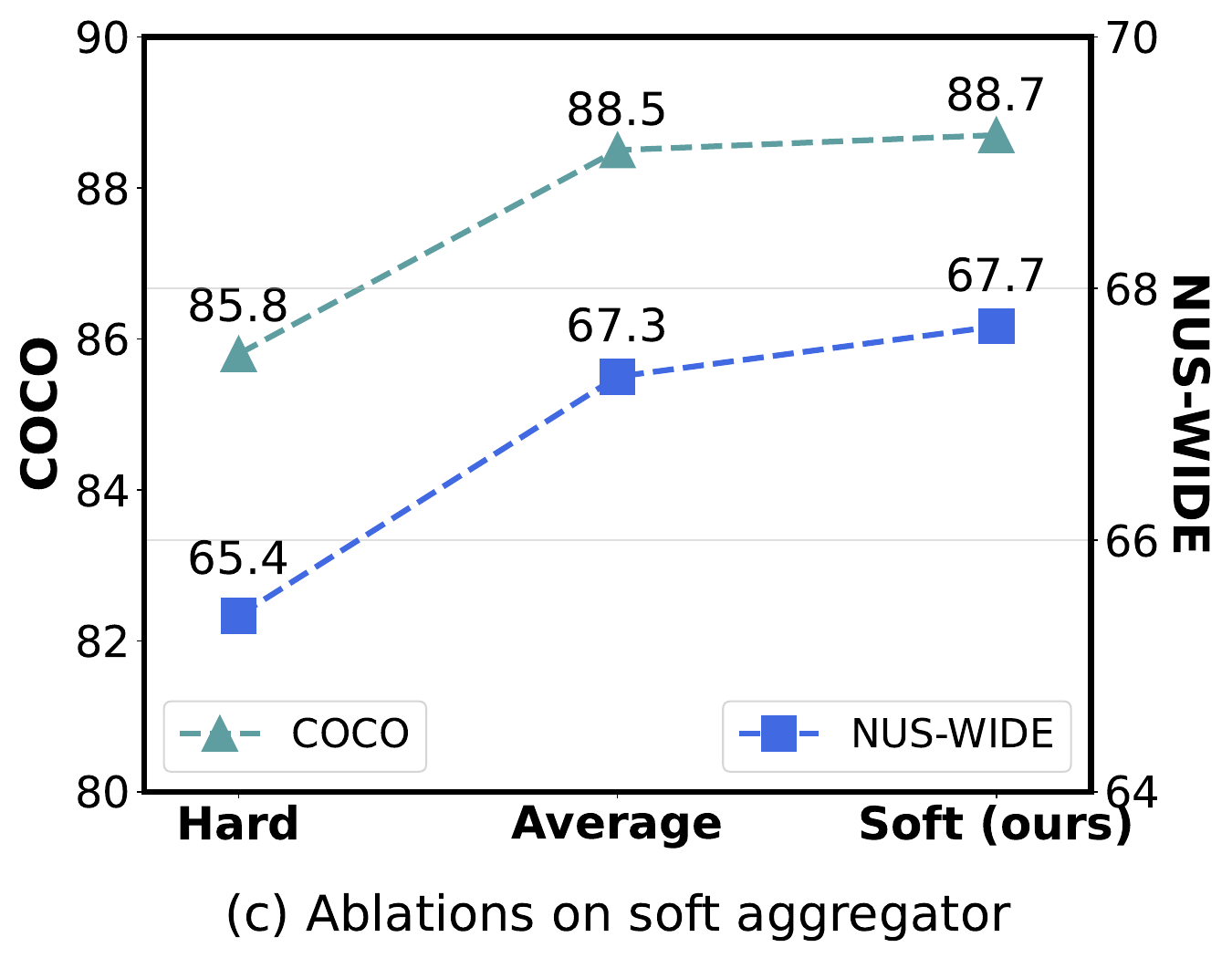}
    \end{minipage}
    \hspace{-3mm}
    \begin{minipage}[t]{0.25\textwidth}
    \centering
    \includegraphics[width=1\linewidth]{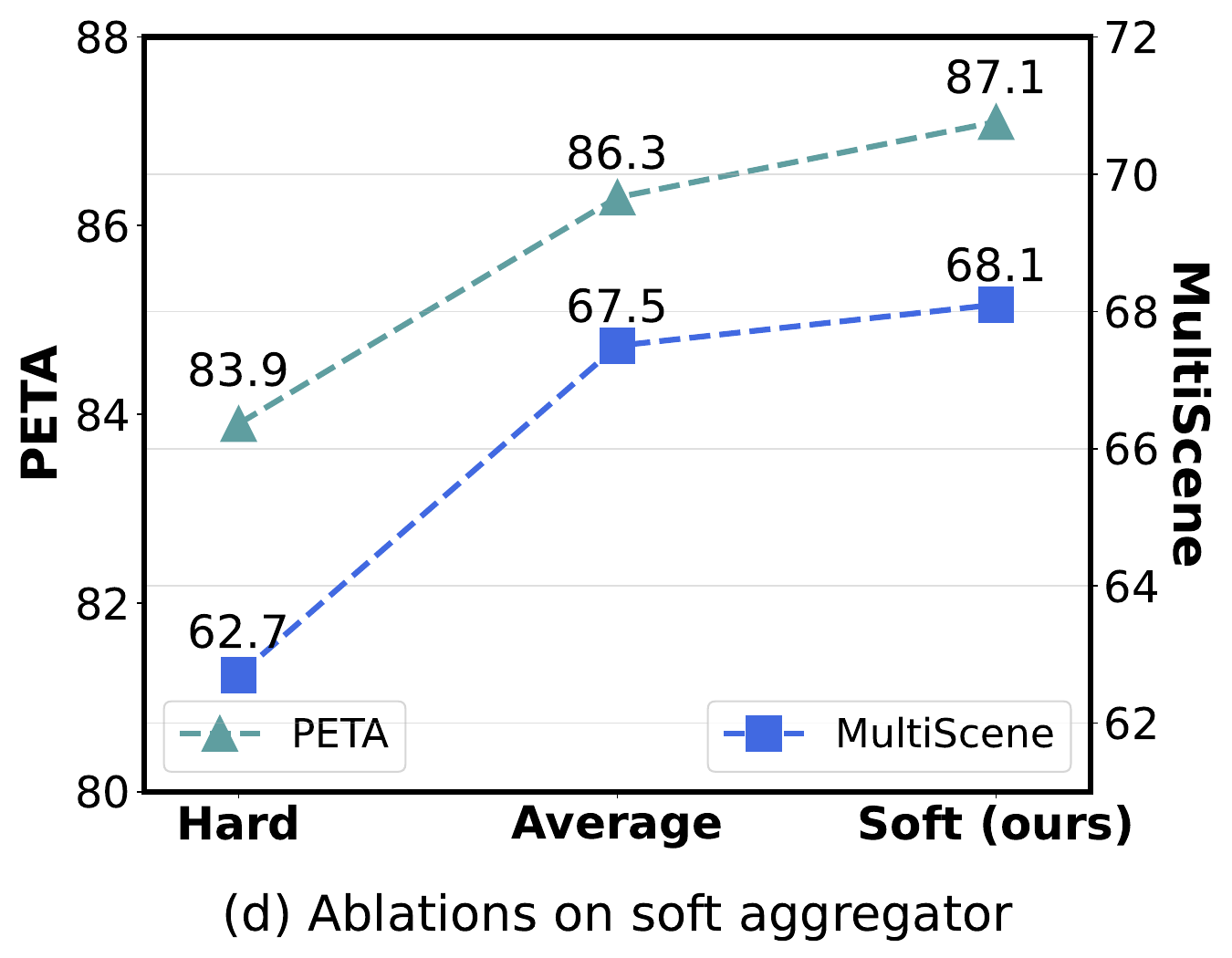}
    \end{minipage}
    \caption{Ablation study (\%) on the global-regional framework and the soft aggregator in regional branch. ``G+R'' denotes the framework using both global and regional branches. ``Hard'' denotes using a hard aggregator and ``Average'' means simply averaging the results from different regions.}
    \label{fig:abl_agg}
\end{figure*}

\begin{table}
\caption{Comparison (\%) to state-of-the-art methods on NUS-WIDE. \dag~indicates ViT-B/16 backbone is used.}
    \label{tab:main_nus}
    \centering
    \scalebox{0.92}{
        \begin{tabular}{p{56pt}<{\raggedright}|p{25pt}<{\centering}p{25pt}<{\centering}p{25pt}<{\centering}|p{25pt}<{\centering}p{25pt}<{\centering}}
        \multirow{2}{*}{\hspace{-2pt}Method} & \multirow{2}{*}{mAP} & \multicolumn{2}{c|}{ALL} & \multicolumn{2}{c}{Top-3}\\
         & & CF1 & OF1 & CF1 & OF1 \\
        \cline{1-6}
        \hspace{-2pt}\rule{0pt}{7pt}CMA~\cite{you2020cross} & 61.4 & 60.5 & 73.7 & 55.5 & 70.0\\
        \hspace{-2pt}GM-MLIC~\cite{wu2021gm} & 62.2 & 61.0 & 74.1 & 55.3 & \textbf{72.5}\\
        \hspace{-2pt}ICME~\cite{chen2019multi2} & 62.8 & 60.7 & 74.1 & 56.3 & 70.6\\
        \hspace{-2pt}ASL~\cite{ridnik2021asymmetric} & 63.9 & 62.7 & 74.6 & - & - \\
        \hspace{-2pt}ML-Decoder~\cite{ridnik2023ml} & 64.6 & - & - & - & - \\
        \hspace{-2pt}DualCoOp~\cite{sun2022dualcoop} & 64.6 & - & - & - & - \\
        \hspace{-2pt}SALGL~\cite{zhu2023scene} & 66.3 & 64.1 & 75.4 & 59.5 & 71.0\\
        \rowcolor{Gray} \hspace{-2pt}\textbf{SSPA} & \textbf{67.7} & \textbf{65.0} & \textbf{75.5} & \textbf{60.7} & 71.2\\
        \cline{1-6}
        \hspace{-2pt}\rule{0pt}{7pt}Q2L-TRL~\cite{liu2021query2label} & 66.3 & 64.0 & 75.0 & - & - \\
        \hspace{-2pt}PatchCT$^{\dag}$~\cite{li2023patchct} & 68.1 & 65.5 & 74.7 & 61.2 & 71.0\\
        \rowcolor{Gray} \hspace{-2pt}\textbf{SSPA}$^{\dag}$ & \textbf{69.9} & \textbf{66.9} & \textbf{76.0} & \textbf{61.9} & \textbf{71.9} \\
        \end{tabular}
    }
\end{table}

\subsection{Ablation Studies}
\noindent \textbf{Effect of global-regional framework and soft aggregator.}
As shown in Fig.~\ref{fig:abl_agg}, incorporating regional cues largely enhances the performance while aggregating global and regional decisions brings further improvements (3.3\%/3.4\% mAP on COCO/NUS).
For regional branch, our proposed soft aggregator outperforms the simple averaging approach.
The performance gap of hard aggregator~\cite{abdelfattah2023cdul} is due to that rigid rules are error-prone, and a softer version mitigates the problem.

\begin{figure}[t]
    \centering
    \begin{minipage}[t]{0.24\textwidth}
    \centering
    \includegraphics[width=1\linewidth]{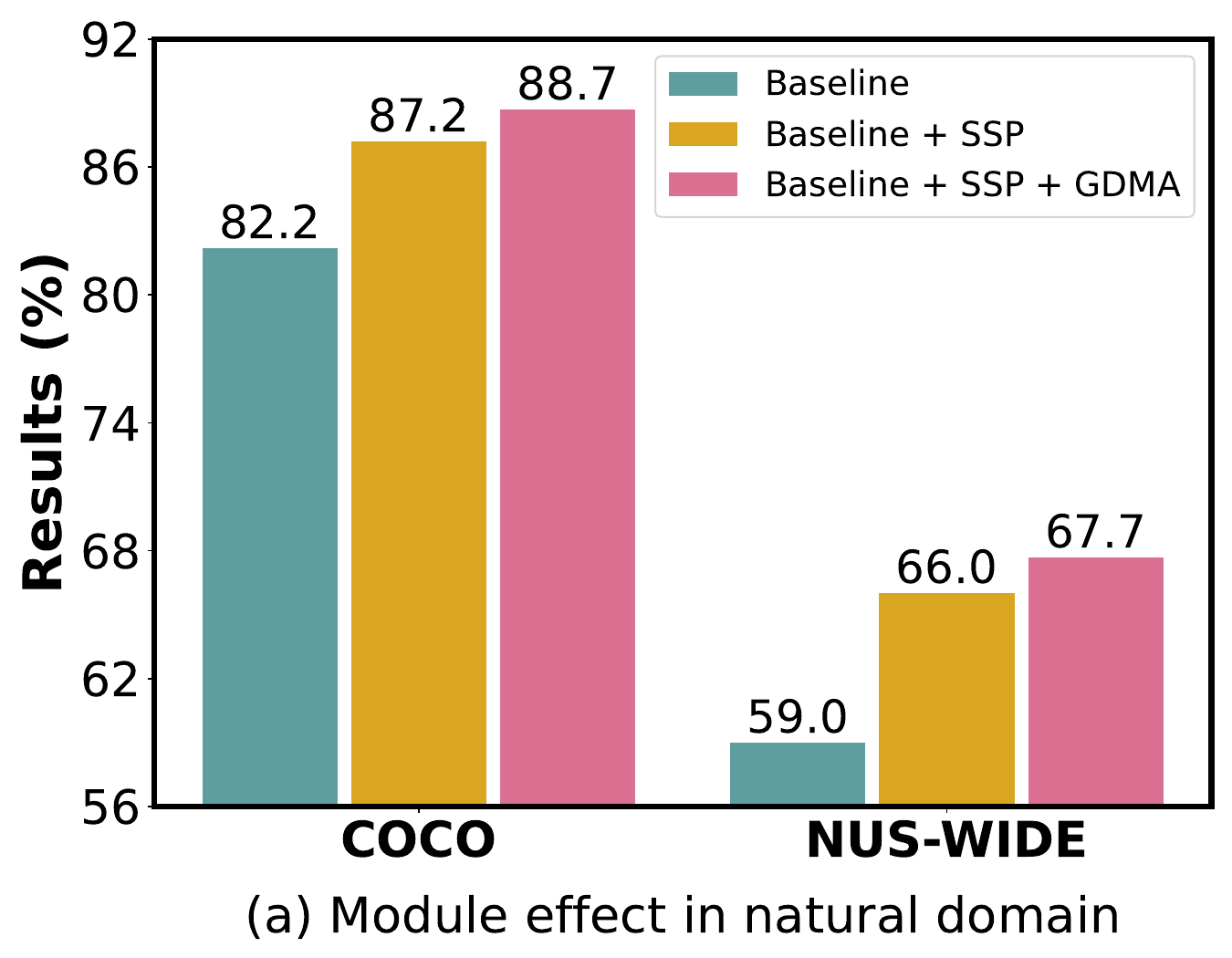}
    \end{minipage}
    \hspace{-3mm}
    \begin{minipage}[t]{0.24\textwidth}
    \centering
    \includegraphics[width=1\linewidth]{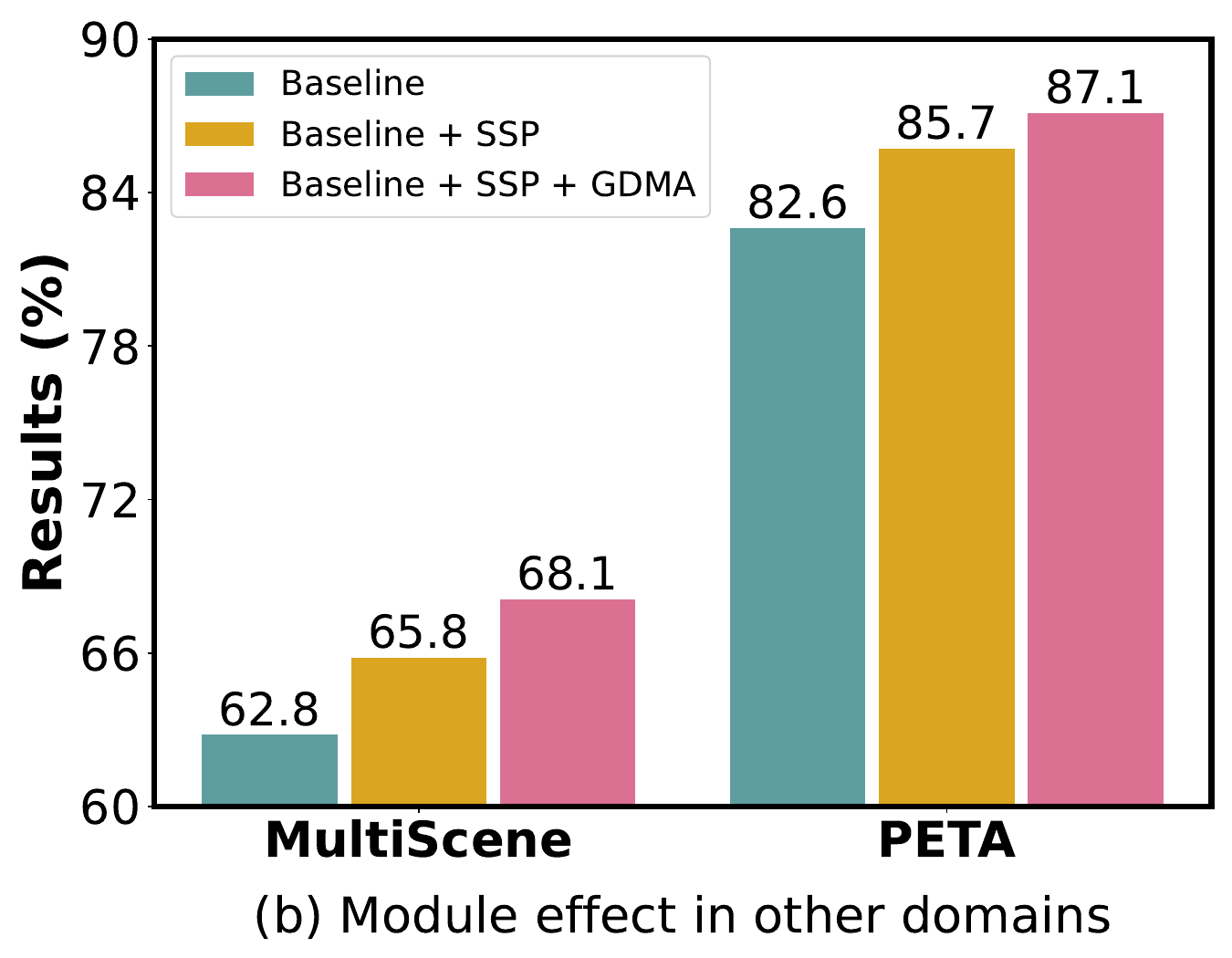}
    \end{minipage}
    \caption{Ablation study (\%) on the proposed SSP and GDMA. ``Baseline'' denotes the fine-tuned CLIP model with text template ``A photo of a \{category\}''.}
    \label{fig:abl0}
\end{figure}

\noindent \textbf{Effect of proposed modules.}
Fig.~\ref{fig:abl0} shows the effectiveness of the proposed SSP and GDMA.
We employ CLIP with the fixed template ``A photo of a \{category\}'' as baseline.
The limited performance indicates that directly transferring CLIP is not effective for MLR.
In contrast, SSP enhances mAP by a significant margin (5.0\%/7.0\% mAP on COCO/NUS), showing the effectiveness of our prompting pipeline. 
While GDMA further improves the performance, which demonstrates that both SSP and GDMA play pivotal roles in SSPA.

\noindent\textbf{Ablation on the split stage of SSP.}
As shown in Table~\ref{tab:abl1}, the performance degrades without KAP or CAP, indicating the importance of split prompting.
Removing LLM descriptions also results in a performance drop, which indicates that our method benefits from the inherent knowledge within LLMs.
Besides, DSF brings further promotion, which is due to that incorporating visual context facilitates the learning of downstream semantics.
Notably, on PETA and MultiScene, CAP (+1.9\%/+1.7\%) is more effective than KAP (+1.0\%/+0.9\%), which indicates that learning of downstream semantics is more important when transferring to those domains of a large gap.

\begin{table*}[t]
    \caption{Comparison (\%) to state-of-the-art methods on pedestrian attribute recognition. We evaluate on three widely-used datasets, i.e., PA100K~\cite{liu2017hydraplus}, RAPv1~\cite{li2018richly} and PETA~\cite{deng2014pedestrian}. Primary metrics (i.e., mA and F1) are highlighted in \textcolor{red}{\textbf{red}}. ``ENet'' denotes EfficientNet.}
    \label{tab:main_par}
    \centering
    \scalebox{0.9}{
        \begin{tabular}{p{65pt}<{\raggedright}p{36pt}<{\centering}p{30pt}<{\centering}|p{14pt}<{\centering}p{15pt}<{\centering}p{15pt}<{\centering}p{15pt}<{\centering}p{15pt}<{\centering}|p{14pt}<{\centering}p{15pt}<{\centering}p{15pt}<{\centering}p{15pt}<{\centering}p{15pt}<{\centering}|p{14pt}<{\centering}p{15pt}<{\centering}p{15pt}<{\centering}p{15pt}<{\centering}p{15pt}<{\centering}}
        \multirow{2}{*}{\hspace{-2pt}Method} & \multirow{2}{*}{Reference} & \multirow{2}{*}{Backbone} & \multicolumn{5}{c|}{PA100K} & \multicolumn{5}{c|}{RAPv1} & \multicolumn{5}{c}{PETA}\\
        \cline{4-18}
        & & & \rule{0pt}{7pt}mA & Acc & Prec & Recall & F1 & mA & Acc & Prec & Recall & F1 & mA & Acc & Prec & Recall & F1 \\
        \cline{1-18}
        \hspace{-2pt}\rule{0pt}{7pt}VRKD~\cite{li2019pedestrian} & IJCAI 19 & ResNet50 & 77.9 & 78.5 & 88.4 & 86.1 & 87.2 & 78.3 & 69.8 & \textbf{82.1} & 80.4 & 81.2 & 84.9 & 81.0 & 88.4 & 87.5 & 87.9 \\
        \hspace{-2pt}JLAC~\cite{tan2020relation} & AAAI 20 & ResNet50 & 82.3 & 79.5 & 87.5 & 87.8 & 87.6 & 83.7 & 69.2 & 79.3 & 82.4 & 80.8 & 87.0 & 80.4 & 87.8 & 87.1 & 87.5 \\
        \hspace{-2pt}SSC$_{soft}$~\cite{jia2021spatial} & ICCV 21 & ResNet50 &81.9 & 78.9 & 86.0 & 89.1 & 86.9 & 82.8 & 68.4 & 75.1 & \textbf{87.5} & 80.4 & 86.5 & 79.0 & 86.0 & 87.1 & 87.0 \\
        \hspace{-2pt}VAC-Combine~\cite{guo2022visual} & IJCV 22 & ResNet50 &82.2 & 80.7 & \textbf{88.7} & 88.1 & 88.4 & 81.3 & 70.1 & 81.6 & 81.5 & 81.5 & - & - & - & - & - \\
        \hspace{-2pt}EALC~\cite{weng2023exploring} & NC 23 & ENet-B4 & 81.5 & 80.3 & 87.3 & 89.0 & 88.1 & 83.3 & 69.7 & 79.8 & 83.6 & 81.7 & 86.8 & \textbf{81.7} & \textbf{88.6} & 88.2 & 88.4 \\
        \hspace{-2pt}DAFL~\cite{jia2022learning} & AAAI 22 & ResNet50 & 83.5 & 80.1 & 87.0 & 89.2 & 88.1 & 83.7 & 68.2 & 77.4 & 83.4 & 80.3 & 87.0 & 78.9 & 85.8 & 87.0 & 86.4 \\
        \hspace{-2pt}CAS-SAL-FR~\cite{yang2021cascaded} & IJCV 22 & ResNet50 &82.9 & 79.6 & 86.8 & 87.8 & 85.2 & \textcolor{red}{\textbf{84.2}} & 68.6 & 77.6 & 83.8 & 80.6 & 86.4 & 79.9 & 87.0 & 87.3 & 87.2 \\
        \hspace{-2pt}CLIP-FT~\cite{radford2021learning} & ICML 21 & ResNet50 & 81.8 & 80.8 & 86.8 & 87.6 & 87.2 & 80.4 & 70.5 & 78.3 & 82.7 & 80.4 & 86.0 & 79.8 & 85.8 & 87.3 & 86.5\\
        \rowcolor{Gray}\hspace{-2pt}\textbf{SSPA} & - & ResNet50 & \textcolor{red}{\textbf{83.7}} & \textbf{81.3} & 87.6 & \textbf{89.7} & \textcolor{red}{\textbf{88.6}} & 83.9 & \textbf{71.1} & 79.0 & 86.1 & \textcolor{red}{\textbf{82.4}} & \textcolor{red}{\textbf{87.1}} & 81.1 & 87.3 & \textbf{90.2} & \textcolor{red}{\textbf{88.7}}\\
        \cline{1-18}
        \hspace{-2pt}\rule{0pt}{7pt}VTB~\cite{cheng2022simple} & TCSVT 22 & ViT-B/16 &83.7 & 80.9 & 87.9 & 89.3 & 88.2 & 82.7 & 69.4 & 78.3 & 84.4 & 80.8 & 85.3 & 79.6 & 86.8 & 87.2 & 86.7 \\
        \hspace{-2pt}DRFormer~\cite{tang2022drformer} & NC 22 & ViT-B/16 &82.5 & 80.3 & 87.6 & 88.5 & 88.0 & 81.8 & 70.6 & 80.1 & 82.8 & 81.4 & \textcolor{red}{\textbf{90.0}} & 81.3 & 85.7 & 91.1 & 88.3 \\
        \hspace{-2pt}PARFormer~\cite{fan2023parformer} & TCSVT 23 & Swin-B &84.0 & 80.3 & 87.5 & 91.1 & 87.7 & 83.8 & 69.7 & 79.2 & \textbf{87.8} & 81.2 & 88.7 & 82.3 & 87.9 & \textbf{91.6} & 88.7 \\
        \hspace{-2pt}ViT-RE~\cite{tan2024vision} & TMM 24 & ViT-B/16 & 84.3 & 81.5 & \textbf{89.8} & 88.0 & 88.9 & \textcolor{red}{\textbf{84.9}} & 69.5 & \textbf{81.2} & 80.8 & 81.0 & 88.2 & 81.6 & \textbf{88.6} & 88.8 & 88.7\\
        \hspace{-2pt}SOFAFormer~\cite{wu2024selective} & AAAI 24 & ViT-B/16 & 83.4 & 81.1 & 88.4 & 89.0 & 88.3 & 83.4 & 70.0 & 80.0 & 83.0 & 81.2 & 87.1 & 81.1 & 87.8 & 88.4 & 87.8 \\
        \hspace{-2pt}CLIP-FT~\cite{radford2021learning} & ICML 21 & ViT-B/16 & 83.9 & 81.5 & 87.4 & 89.4 & 88.4 & 82.8 & 71.9 & 79.1 & 84.7 & 81.8 & 86.6 & 81.3 & 87.1 & 90.0 & 88.5\\
        \rowcolor{Gray}\hspace{-2pt}\textbf{SSPA} & - & ViT-B/16 & \textcolor{red}{\textbf{85.1}} & \textbf{83.3} & 89.1 & \textbf{91.7} & \textcolor{red}{\textbf{90.2}} & 84.1 & \textbf{72.5} & 80.2 & 87.3 & \textcolor{red}{\textbf{83.6}} & 88.9 & \textbf{83.8} & 88.1 & 91.4 & \textcolor{red}{\textbf{89.7}}\\
        \cline{1-18}
        \hspace{-2pt}\rule{0pt}{7pt}PARFormer~\cite{fan2023parformer} & TCSVT 23 & Swin-L & 84.5 & 81.1 & 88.1 & 91.7 & 88.5 & 84.1 & 69.9 & 79.6 & 88.2 & 81.4 & 89.3 & 82.9 & 88.1 & 92.0 & 89.1 \\
        \hspace{-2pt}PromptPAR~\cite{wang2023pedestrian} & - & ViT-L/14 & \textcolor{red}{\textbf{87.5}} & 83.8 & 89.3 & 91.7 & 90.2 & 85.5 & 71.6 & 79.6 & 86.1 & 82.4 & 88.8 & 82.8 & 89.0 & 89.7 & 89.2 \\
        \hspace{-2pt}CLIP-FT~\cite{radford2021learning} & ICML 21 & ViT-L/14 & 85.0 & 83.6 & 88.6 & 91.3 & 89.9 & 83.9 & 71.1 & 79.4 & 87.1 & 83.1 & 88.3 & 82.5 & 87.9 & 90.3 & 89.0\\
        \rowcolor{Gray}\hspace{-2pt}\textbf{SSPA} & - & ViT-L/14 & 87.0 & \textbf{84.9} & \textbf{89.6} & \textbf{93.0} & \textcolor{red}{\textbf{91.3}} & \textcolor{red}{\textbf{85.9}} & \textbf{73.3} & \textbf{80.2} & \textbf{88.1} & \textcolor{red}{\textbf{84.0}} & \textcolor{red}{\textbf{90.2}} & \textbf{84.7} & \textbf{89.2} & \textbf{92.1} & \textcolor{red}{\textbf{90.6}}\\
        \end{tabular}
    }
\end{table*}

\begin{table*}[t]
    \caption{Comparison (\%) to state-of-the-art methods on remote sensing datasets. We evaluate on three popular datasets, including MultiScene~\cite{hua2021multiscene}, MLRSNet~\cite{qi2020mlrsnet} and AID~\cite{xia2017aid}.}
    \label{tab:main_rs}
    \centering
    \scalebox{0.9}{
        \begin{tabular}{p{68pt}<{\raggedright}p{32pt}<{\centering}p{31pt}<{\centering}|p{14pt}<{\centering}p{15pt}<{\centering}p{15pt}<{\centering}p{15pt}<{\centering}p{15pt}<{\centering}|p{14pt}<{\centering}p{15pt}<{\centering}p{15pt}<{\centering}p{15pt}<{\centering}p{15pt}<{\centering}|p{14pt}<{\centering}p{15pt}<{\centering}p{15pt}<{\centering}p{15pt}<{\centering}p{14pt}<{\centering}}
       \multirow{3}{*}{\hspace{-2pt}Method} & \multirow{3}{*}{Reference} & \multirow{3}{*}{Backbone} & \multicolumn{5}{c|}{MultiScene} & \multicolumn{5}{c|}{MLRSNet} & \multicolumn{5}{c}{AID}\\
        \cline{4-18}
        & & & \multirow{2}{*}{mAP} & \multicolumn{2}{c}{\rule{0pt}{8pt}ALL} & \multicolumn{2}{c|}{Top-3} & \multirow{2}{*}{mAP} & \multicolumn{2}{c}{ALL} & \multicolumn{2}{c|}{Top-3} & \multirow{2}{*}{mAP} & \multicolumn{2}{c}{ALL} & \multicolumn{2}{c}{Top-3}\\
        & & & & CF1 & OF1 & CF1 & OF1 & & CF1 & OF1 & CF1 & OF1 & & CF1 & OF1 & CF1 & OF1 \\
        \cline{1-18}
        \hspace{-2pt}\rule{0pt}{8pt}RemoteCLIP-FT~\cite{liu2024remoteclip} & TGRS 24 & ResNet50 & 66.9 & 63.2 & 73.9 & \textbf{52.0} & 65.2 & 97.7 & 90.6 & 92.6 & \textbf{67.9} & 70.1 & 84.4 & \textbf{78.2} & 90.3 & \textbf{49.3} & 65.8\\
        \rowcolor{Gray}\hspace{-2pt}\textbf{SSPA} & - & ResNet50 & \textbf{68.1} & \textbf{65.2} & \textbf{75.0} & \textbf{52.0} & \textbf{66.1} & \textbf{98.1} & \textbf{91.4} & \textbf{93.4} & 65.6 & \textbf{70.3} & \textbf{85.5} & 77.6 & \textbf{91.3} & 46.5 & \textbf{66.2} \\
        \cline{1-18}
        \rule{0pt}{8pt}\hspace{-2pt}RemoteCLIP-FT~\cite{liu2024remoteclip} & TGRS 24 & ViT-B/32 & 67.2 & 63.3 & 73.7 & 52.5 & 65.6 & 97.8 & 91.1 & 93.1 & \textbf{66.7} & 70.2 & 85.1 & 81.7 & 91.5 & \textbf{50.9} & 66.3 \\
        \rowcolor{Gray}\hspace{-2pt}\textbf{SSPA} & - & ViT-B/32 & \textbf{68.5} & \textbf{65.6} & \textbf{75.2} & \textbf{53.9} & \textbf{66.2} & \textbf{98.5} & \textbf{91.5} & \textbf{93.4} & 65.7 & \textbf{70.6} & \textbf{87.1} & \textbf{81.9} & \textbf{91.7} & 49.3 & \textbf{66.5}\\
        \end{tabular}
    }
\end{table*}

\begin{table}[t]
    \centering
    \caption{Ablation study (\%) on the split stage of SSP.}
    \label{tab:abl1}
		\centering
        \scalebox{0.93}{
        \begin{tabular}{p{18pt}<{\centering}p{28pt}<{\centering}|p{18pt}<{\centering}p{18pt}<{\centering}|p{18pt}<{\centering}p{18pt}<{\centering}p{18pt}<{\centering}p{30pt}<{\centering}}
        \multicolumn{2}{c|}{KAP} & \multicolumn{2}{c|}{CAP} & \multirow{2}{*}{COCO} & \multirow{2}{*}{NUS} & \multirow{2}{*}{PETA} & \multirow{2}{*}{MultiScene} \\
        LLM & Template & Soft & DSF & & & &\\
        \cline{1-8}
        \rule{0pt}{7pt}\ding{51} & \ding{51} & & & 87.5 & 66.6 & 85.2 & 66.4  \\
        & & \ding{51} & \ding{51} & 87.7 & 66.4 & 86.1 & 67.2  \\
        & \ding{51} & \ding{51} & \ding{51} & 88.2 & 67.2 & 86.5 & 67.7 \\
        \ding{51} & \ding{51} & \ding{51} & & 87.9 & 67.0 & 86.3 & 67.4 \\
        \rowcolor{Gray} \ding{51} & \ding{51} & \ding{51} & \ding{51} & 88.7 & 67.7 & 87.1 & 68.1 \\
        \end{tabular}
    }
\end{table}

\begin{table}[t]
    \centering
    \caption{Ablation study (\%) on synthesis stage of SSP.}
    \label{tab:abl3}
        \scalebox{0.93}{
        \begin{tabular}{p{70pt}<{\centering}|p{25pt}<{\centering}p{25pt}<{\centering}p{25pt}<{\centering}p{30pt}<{\centering}}
        Method & COCO & NUS & PETA & MultiScene\\
        \cline{1-5}
        \rule{0pt}{7pt}Summation & 87.8 & 66.5 & 86.1 & 67.3\\
        Concatenation & 87.9 & 66.7 & 86.4 & 67.4\\
        MLP & 88.2 & 67.1 & 86.8 & 67.7 \\
        \rowcolor{Gray} QSM & 88.7 & 67.7 & 87.1 & 68.1 \\
        \end{tabular}
    }
\end{table}

\noindent\textbf{Ablation on the synthesis stage of SSP.}
As shown in Table~\ref{tab:abl3}, 
direct summation or concatenation yields degraded performance, implying that the generic knowledge and downstream semantics require more fine-grained synthesis.
Compared to MLP synthesis, which shares similar number of parameters and computation overhead with our proposed QSM, the QSM achieves better performance, revealing the superior capabilities of quaternion latent space to capture comprehensive relationships and inter-modal dependencies.

\noindent\textbf{Ablation on GDMA.}
In Table~\ref{tab:abl4}, we show that employing visual-to-semantic attention (V-to-S) or semantic-to-visual attention (S-to-V) individually leads to degraded performance, while the latter yields worse results.
Overall, using both concurrently (i.e., GDMA) brings significant improvements. 
This confirms the superiority of bidirectional interactions and further indicates that incorporating visual context into label representations is of more importance. 
One reason is that V-to-S attention enables input-adaptive category centers during inference.
As for the gate mechanism in GDMA, 
Table~\ref{tab:abl2} further shows that the proposed gated alignment brings notable improvements compared to direct interaction.

\begin{table}[t]
    \centering
    \caption{Ablation study (\%) on the GDMA.}
    \label{tab:abl4}
        \scalebox{0.93}{
        \begin{tabular}{p{30pt}<{\centering}p{30pt}<{\centering}|p{30pt}<{\centering}p{30pt}<{\centering}p{30pt}<{\centering}p{30pt}<{\centering}}
        V-to-S & S-to-V & COCO & NUS & PETA & MultiScene \\
        \cline{1-6}
         & & \rule{0pt}{7pt}87.2 & 66.0 & 85.7 & 65.8\\
        \ding{51} & & 88.2 & 67.1 & 86.5 & 67.2\\
        & \ding{51} & 87.8 & 66.5 & 86.3 & 67.5\\
        \rowcolor{Gray} \ding{51} & \ding{51} & 88.7 & 67.7 & 87.1 & 68.1\\
        \end{tabular}
    }
\end{table}

\begin{table}[t]
    \centering
    \caption{Analysis (\%) on the cross-modal interactions.}
    \label{tab:abl2}
        \scalebox{0.93}{
        \begin{tabular}{p{40pt}<{\centering}p{40pt}<{\centering}p{29pt}<{\centering}|p{29pt}<{\centering}p{29pt}<{\centering}}
        Interaction & Attention & Text & COCO & NUS \\
        \cline{1-5}
        \rule{0pt}{7pt}Uni-D & - & CLIP & 84.3 & 64.1\\
        Bi-D & w/o Gate & CLIP & 88.0 & 66.6\\
        \rowcolor{Gray}Bi-D & w/ Gate & CLIP & 88.7 & 67.7\\
        \cline{1-5}
        \rule{0pt}{7pt}Uni-D & - & BERT & 84.8 & 64.8 \\
        Bi-D & w/o Gate & BERT & 87.4 & 66.3\\
        Bi-D & w/ Gate & BERT & 88.4 & 67.2\\
        \end{tabular}
    }
\end{table}

\subsection{Further Analyses}
\label{sec:further}
\begin{figure*}[t]
    \centering
    \includegraphics[width=0.98\linewidth]{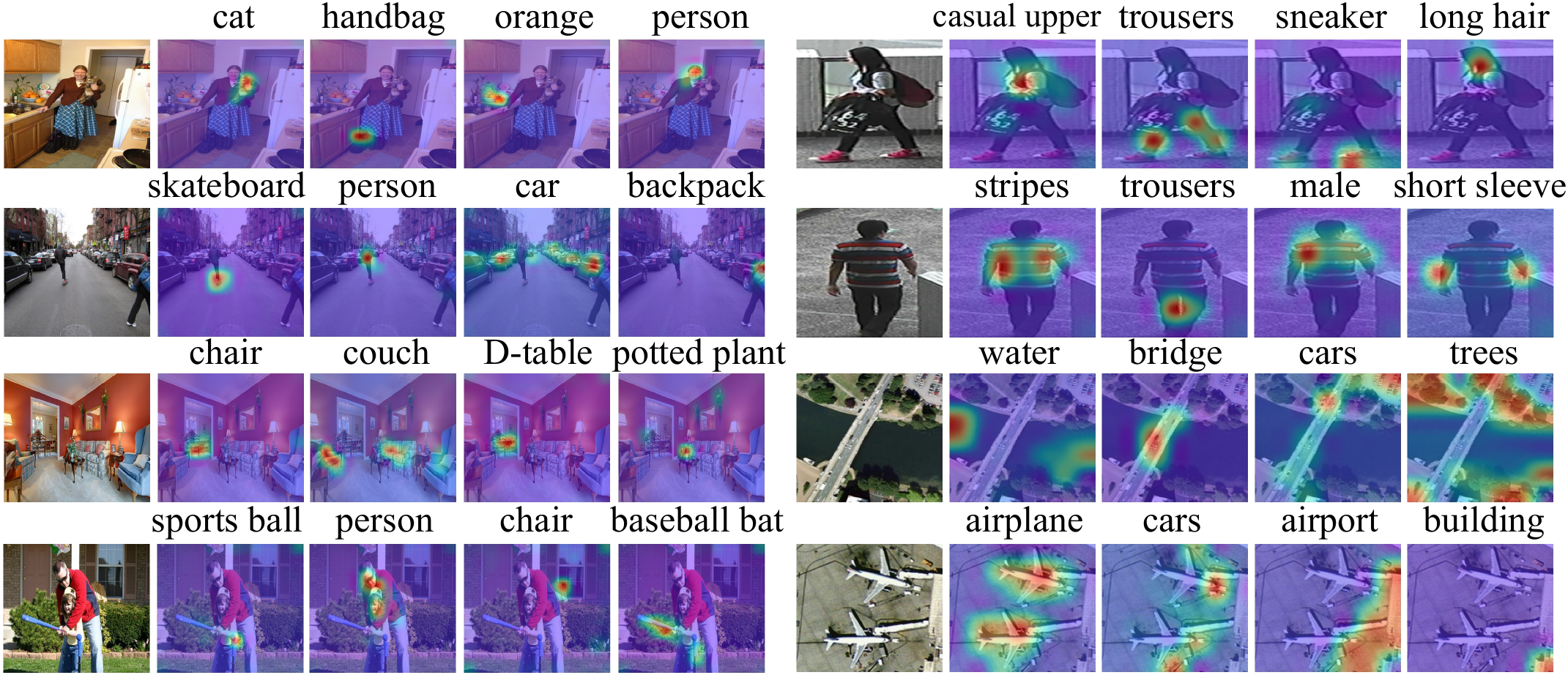}
    \caption{Visualization of the attention maps in GDMA on the test sets of COCO~\cite{lin2014microsoft}, PETA~\cite{deng2014pedestrian} and MultiScene~\cite{hua2021multiscene}.
    Brighter color means higher attention weight.
    SSPA can precisely perceive target objects of different sizes, similar appearances and specific domain semantics.
    Best viewed in colors.}
	\label{fig:attnmap}
\end{figure*}

\noindent \textbf{Visualization of attention maps.}
As shown in Fig.~\ref{fig:attnmap}, SSPA can accurately perceive 
common objects of distinct size (e.g., ``backpack'' and ``car'')and similar appearance (e.g., ``chair'' and ``couch'') in natural images.
For pedestrian images, SSPA can focus on specific body parts, e.g., emphasizing the arms area when detecting ``short sleeves'', and concentrating on the legs when determining the lower body clothing.
For remote sensing images, SSPA can perceive large land covers (e.g., ``trees'' and ``water'') and fine-grained objects (e.g., ``cars'' and ``airplane'').
These observations demonstrate the excellent discriminative capability of the proposed SSPA.

\begin{figure}[t]
    \centering
    \includegraphics[width=1.\linewidth]{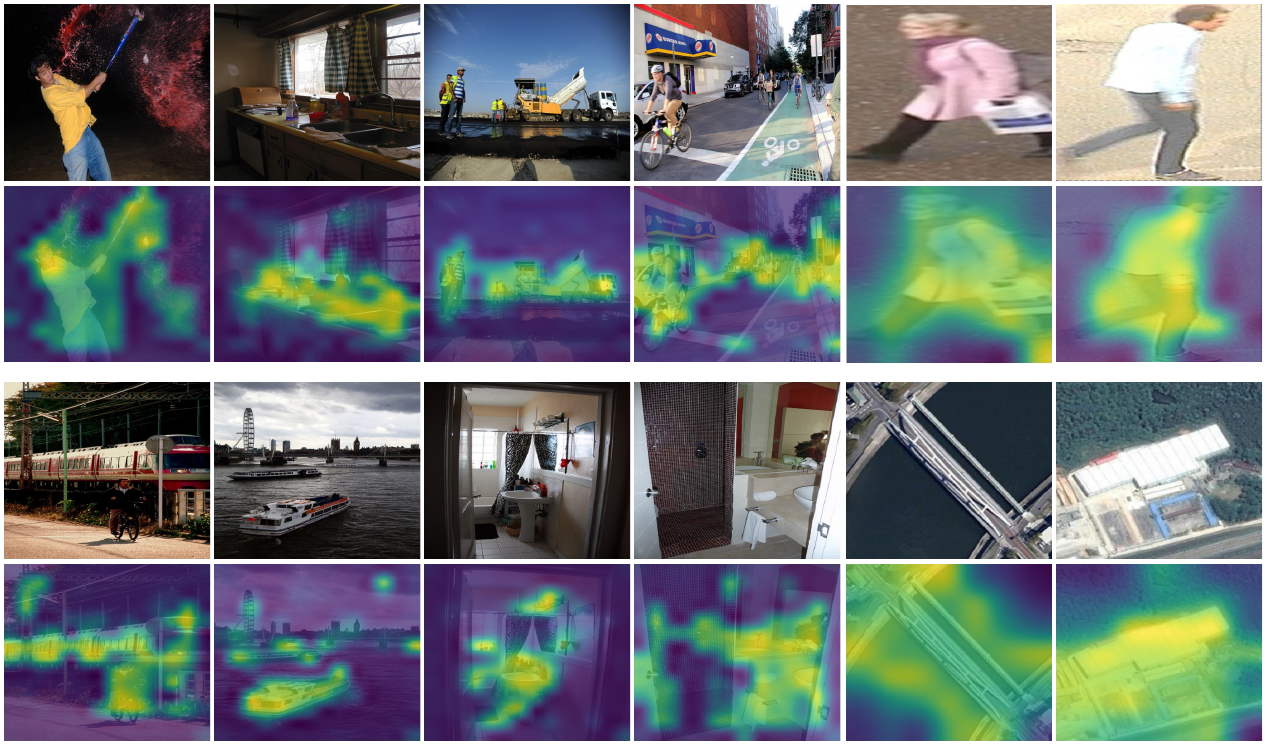}
    \caption{Visualization of gate vectors on the test sets of COCO~\cite{lin2014microsoft}, PETA~\cite{deng2014pedestrian} and MultiScene~\cite{hua2021multiscene}.
    Brighter color means higher passing rates.
    The gate vectors effectively suppress most background areas. Best viewed in colors.}
    \label{fig:gate_vector}
\end{figure}

\noindent\textbf{Interpretability of gate vectors.}
To validate the effectiveness of the proposed gated alignments, we visualize the gate vectors in Fig.~\ref{fig:gate_vector}.
For remote sensing images, the gate vectors activate most areas since the land covers often dominate the whole image.
For natural and pedestrian images, the gate mechanism successfully suppresses most background information, which can enhance the efficiency of cross-modal interactions, yielding more pertinent visual and label representations.

\noindent\textbf{Is using label representations as category centers better?}
In Table~\ref{tab:abl2}, ``Uni-D'' denotes a baseline similar to previous approaches, where $C$ additional classifiers are learned.
Our method improves the mAP by 4.4\% and 3.6\% on COCO and NUS, respectively, suggesting the superiority of constructing dynamic category centers through label representations.
Similar conclusions can be drawn on BERT~\cite{devlin2018bert}, showing the robustness of SSPA to the text encoder.

\begin{table}[t]
    \centering
    \caption{Analysis on the choice of LLM.}
    \label{tab:abl_llm}
        \scalebox{0.93}{
        \begin{tabular}{p{30pt}<{\centering}p{60pt}<{\centering}|p{30pt}<{\centering}p{30pt}<{\centering}}
        Method & LLM & COCO & NUS\\
        \cline{1-4}
        \rule{0pt}{7pt}SSPA & LLaMA-7B & 88.6 & 67.6\\
        SSPA & Vicuna-7B & 88.7 & 67.5\\
        SSPA & LLaMA3-8B & 88.7 & 67.7\\
        \end{tabular}
    }
\end{table}

\begin{table}[t]
    \centering
    \caption{Analysis on the ImageNet pretraining.}
    \label{tab:abl_imagenet}
        \scalebox{0.93}{
        \begin{tabular}{p{42pt}<{\raggedright}|p{34pt}<{\centering}p{44pt}<{\centering}p{35pt}<{\centering}|p{20pt}<{\centering}p{20pt}<{\centering}}
        \hspace{-2pt}Method & Backbone & Pre-train & Resolution & COCO & NUS\\
        \cline{1-6}
        \hspace{-2pt}\rule{0pt}{7pt}ML-GCN~\cite{chen2019multi} & ResNet101 & ImageNet-1K & (448,448) & 83.0 & 62.5\\
        \hspace{-2pt}Q2L~\cite{liu2021query2label} & ResNet101 & ImageNet-1K & (448,448) & 84.9 & 65.0\\
        \rowcolor{Gray}\hspace{-2pt}SSPA & ResNet101 & ImageNet-1K & (448,448) & 86.5 & 66.3\\
        \end{tabular}
    }
\end{table}

\noindent\textbf{Data bias of LLM.}
Different pre-training corpora may lead to biased output from LLM.
In Table~\ref{tab:abl_llm}, we reveal that appropriate prompts are sufficient for LLMs to generate accurate information with less noise, regardless of the pertaining corpora (or data bias).
When replacing LLaMA3-8B with Vicuna or LLaMA (using the same prompt template), similar performance is achieved, exhibiting stableness and robustness.

\noindent\textbf{ImageNet pretraining.}
As shown in Table~\ref{tab:abl_imagenet}, with ImageNet pre-trained backbone, our method still outperforms previous methods (e.g., ML-GCN and Q2L) significantly, which shows the alignment abilities of SSPA.
Besides, we suggest using pre-trained VLMs (e.g., CLIP) for better performance.

\noindent\textbf{Computation overhead.}
\textit{During deployment, the text features from the text encoder can be extracted in advance and 
preserved offline on testing devices, and the corresponding computation could be ignored.}
As shown in Table~\ref{tab:comp},
compared to the baseline, our method achieves considerable performance gains with limited computational increase.
Moreover, SSPA achieves much better performance with comparable FLOPs and parameters than previous SOTAs, e.g., Q2L. 
In fact, the parameters and FLOPs of our QSM are similar to that of MLP, and we only require one QSM and one GDMA layer, which only brings a few extra computations.

\begin{table}[t]
    \centering
    \caption{Analysis on the computation overhead. ``M'' (Million) denotes the number of parameters.}
    \label{tab:comp}
    \scalebox{0.93}{
        \begin{tabular}{p{58pt}<{\raggedright}|p{40pt}<{\centering}p{40pt}<{\centering}|p{20pt}<{\centering}p{20pt}<{\centering}}
        \hspace{-2pt}Method & GFLOPs & Params [M] & COCO & NUS\\
        \cline{1-5}
        \hspace{-2pt}\rule{0pt}{7pt}DualCoOp~\cite{sun2022dualcoop} & 36.6 & 42.5 & 85.3 & 64.6 \\
        \hspace{-2pt}ML-Decoder~\cite{ridnik2023ml} & 37.2 & 47.3 & 86.6 & 64.2 \\
        \hspace{-2pt}Q2L~\cite{liu2021query2label} & \textbf{43.2} & \textbf{143.1} & 84.9 & 65.0 \\
        \cline{1-5}
        \hspace{-2pt}\rule{0pt}{7pt}Baseline & 36.8 & 43.6 & 81.6 & 58.6 \\
        \rowcolor{Gray}\hspace{-2pt}SSPA & 37.7 & 50.8 & \textbf{88.7} & \textbf{67.7} \\
        \end{tabular}
    }
\end{table}

\section{Conclusion}
\label{sec:con}
\noindent In this work, we proposed SSPA, a novel framework for multi-label image recognition.
We developed an SSP strategy to aggregate generic knowledge from LLMs and label semantics from target tasks, which makes great use of linguistic knowledge to enhance the recognition of multiple labels.
We further presented GDMA to bidirectionally interact text embeddings and visual features while filtering out redundant signals, largely enhancing the efficiency of cross-modal alignments.
Extensive experiments on different domains show the state-of-the-art performance and generalization abilities of SSPA.

\noindent\textbf{Limitations and broader impacts.}
The major limitation is that there are unannotated objects in our training images, which could impact the performance in real-world applications. 
One of our future works will focus on overcoming this problem by expanding the method to open-vocabulary scenarios.
As for the social impacts, since this work aims to develop a general method without targeting specific applications, which does not directly involve societal issues.
However, when applied to the pedestrian domain, it can also have social and cultural implications by influencing perceptions of privacy, surveillance, and individual autonomy. 
The ethical and legal implications on the collection, storage, and use of pedestrian attribute data should be carefully considered.

\bibliographystyle{IEEEtran}
\bibliography{IEEEabrv,main}

\end{document}